\pdfoutput=1
\documentclass[11pt]{article}

\usepackage[]{acl}

\usepackage{times}
\usepackage{latexsym}
\usepackage{amsmath}
\usepackage{stmaryrd}
\usepackage{amsfonts}
\usepackage{multirow}
\usepackage{booktabs}
\usepackage{adjustbox}
\usepackage{array}
\usepackage{graphicx} % DO NOT CHANGE THIS
\usepackage{xcolor}
\usepackage{makecell}
\usepackage{arydshln}
\usepackage{textcomp}
\usepackage{longtable}

\definecolor{cyan}{rgb}{0.32, 0.64, 0.80}
\definecolor{orange}{rgb}{0.93, 0.46, 0.14}
\definecolor{green}{rgb}{0.49, 0.70, 0.35}
\definecolor{blue}{rgb}{0.22, 0.44, 0.80}
\definecolor{lightblue}{HTML}{99CCFF}
\definecolor{lightorange}{rgb}{1, 0.8, 0.6}
\definecolor{lightgreen}{HTML}{6BD500}
\definecolor{red}{rgb}{0.99, 0.02, 0.02}
\definecolor{lightred}{HTML}{FF9999}
\definecolor{purple}{rgb}{0.60, 0.41, 0.61}
\definecolor{train}{rgb}{0.98, 0.92, 0.88}
\definecolor{test}{rgb}{0.88, 0.96, 1.0}
\usepackage[T1]{fontenc}
\usepackage[utf8]{inputenc}

\usepackage{microtype}

\NewDocumentCommand{\heng}
{ mO{} }{\textcolor{red}{\textsuperscript{\textit{Heng}}\textsf{\textbf{\small[#1]}}}}

\NewDocumentCommand{\jeongh}
{ mO{} }{\textcolor{orange}{\textsuperscript{\textit{Jeonghwan}}\textsf{\textbf{\small[#1]}}}}

\title{Finer: Investigating and Enhancing Fine-Grained\\Visual Concept Recognition in Large Vision Language Models}

\author{Jeonghwan Kim\quad\quad Heng Ji \\
University of Illinois Urbana-Champaign \\
\texttt{\textmd{\{jk100, hengji\}@illinois.edu}}
}

\begin{document}
\maketitle
\begin{abstract}
Recent advances in instruction-tuned Large Vision-Language Models (LVLMs) have imbued the models with the ability to generate high-level, image-grounded explanations with ease. While such capability is largely attributed to the rich world knowledge contained within the Large Language Models (LLMs), our work reveals their shortcomings in fine-grained visual categorization (FGVC) across six different benchmark settings. Most recent state-of-the-art LVLMs such as LLaVa-1.5, InstructBLIP and GPT-4V not only severely deteriorate in terms of classification performance, e.g., average drop of 65.58 in EM for Stanford Dogs for LLaVA-1.5, but also struggle to generate descriptive visual attributes based on a concept that appears within an input image despite their prominent zero-shot image captioning ability. In-depth analyses show that instruction-tuned LVLMs suffer from modality gap, showing discrepancy when given textual and visual inputs that correspond to the same concept. In an effort to further the community's endeavor in this direction, we propose a multiple granularity attribute-centric benchmark and training mixture, \textbf{\textsc{Finer}}, which aims to establish a ground to evaluate LVLMs' fine-grained visual comprehension ability and provide significantly improved explainability.
% \footnote{We release the dataset and code at: \url{https://github.com/wjdghks950/Finer}}
%\heng{add that in addition to deep analysis you are also offering potential solutions which already showed significant improvement. Also emphasize that in addition to accuracy improvement, your method also significantly enhanced explanability}
\end{abstract}

\section{Introduction}

\begin{figure}[t]
    \centering
    \includegraphics[scale=0.33]{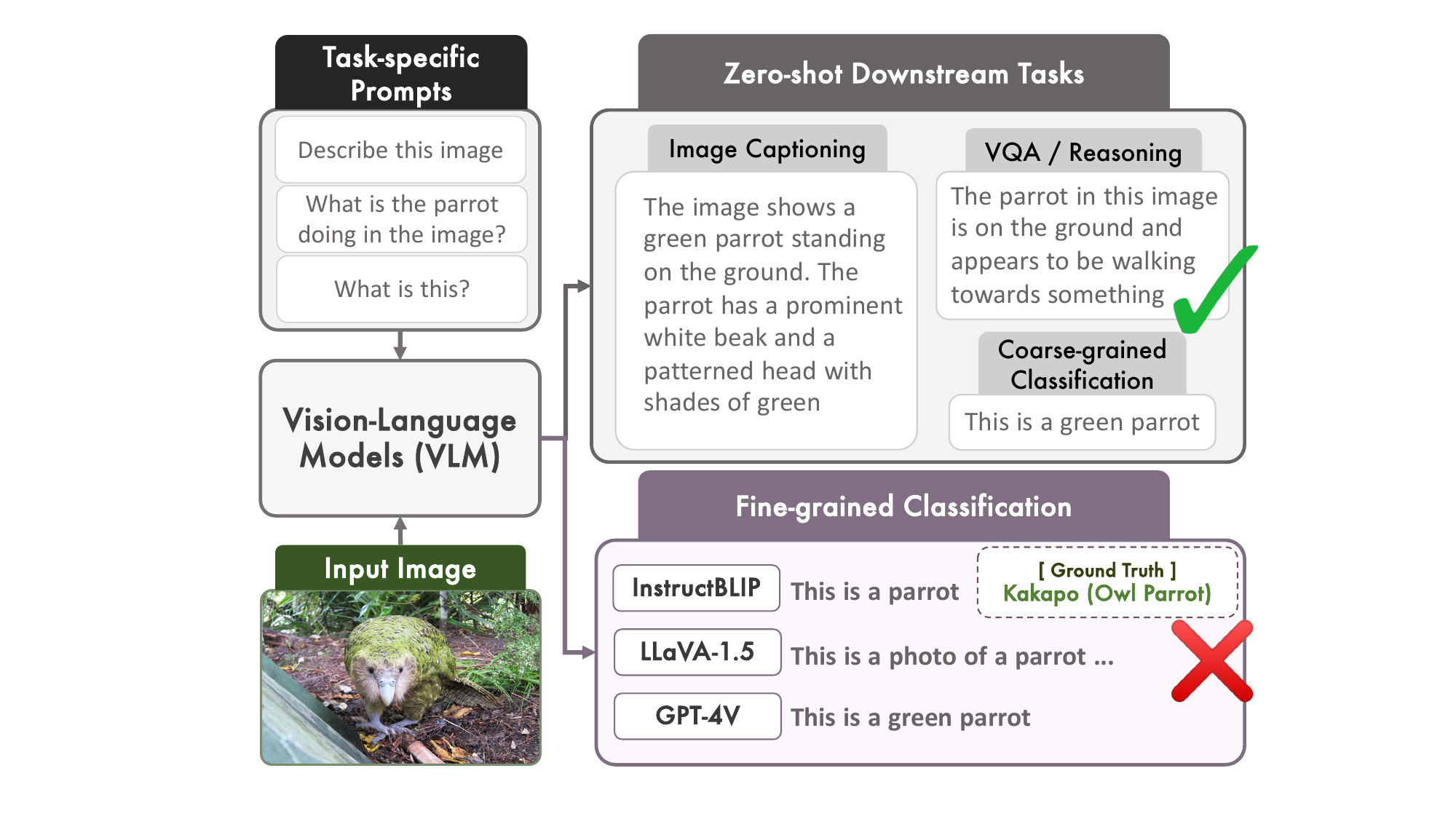}
    \caption{Current state-of-the-art LVLMs exhibit strong zero-shot downstream task solving abilities (e.g., image captioning, VQA, reasoning). However, when prompted to classify the fine-grained concepts, most of them fail to distinguish them into finer categories. Fine-grained classification prompt here is omitted for brevity.}
    \label{fig:overview_figure1}
\end{figure}

In recent years, Large Vision-Language Models (LVLMs) that are able to generate image-grounded text have seen significant progress. Models such as InstructBLIP \citep{dai2023instructblip} and LLaVA \citep{liu2023llava, liu2023improvedllava} have consistently exhibited strong zero-shot capability in generating image captions, visual reasoning and textual descriptions, and even leveraging external knowledge for complex question answering tasks \cite{marino2019ok, schwenk2022okvqa}. Such results across diverse benchmarks indicate that these models, most of them being built on large language models (LLMs) like Vicuna \citep{vicuna2023}, Flan-T5 \citep{chung2022scaling}, Llama \citep{touvron2023llama}, are already equipped with the ability to simultaneously leverage the interplay between the textual parametric knowledge acquired during pre-training and the image understanding ability acquired during instruction-tuning. Notably, all of these models exhibit strong \textit{zero-shot task transferability} to multiple downstream tasks.

Conventionally, in the computer vision domain, many previous works on fine-grained visual classification (FGVC) \citep{fine_image_survey_2022, diao2022metaformer, zhu2022dual, yang2022fine, wang2022knowledgemining} sought to accurately classify diverse images ranging from different types of birds, plants, animals \cite{van2015building, van2018inaturalist} and artificial objects such as cars \cite{krause2013collecting} and aircrafts \cite{maji13fine-grained}. In this work, we investigate whether state-of-the-art LVLMs can combine their image understanding ability and rich textual knowledge acquired during pre-training to handle zero-shot FGVC. To our surprise, while the LVLMs perform almost perfectly, e.g., 98.43 for LLaVA-1.5 (13B) on iNaturalist, at superordinate-level granularity (e.g., \textit{birds}, \textit{jets}), their classification abilities do not extend to the coarse and finer-grained concepts (e.g., \textit{bald eagle}, \textit{F-22 Raptor}), exhibiting substantially deteriorated classification performance (\S \ref{sec:brittleness_vlms}); 46.91 for coarse-level and 1.56 for fine-level categories on iNaturalist. Our empirical analyses of these models reveal that these models suffer from \textit{modality gap}. We empirically demonstrate that such discrepancy stems from LVLMs' limited ability to exploit the rich parametric knowledge given image input, to infer fine-grained concepts. We also show that such constraints lead to diminished fine-grained understanding of the image, preventing these models from generating accurate and detailed visual attributes of the concepts that appear within an image.

We also present an attribute-centric and multiple granularity classification benchmark and training mixture, \textbf{\textsc{Finer}}. Our benchmark constructs concept-indicative attributes for six conventional FGVC benchmarks like iNaturalist \cite{van2018inaturalist} and FGVC-Aircrafts \cite{maji13fine-grained} by (i) generating multiple granular concept labels for visual concept recognition, and (ii) constructing a set of visual attributes per fine-grained concept to measure the ability of LVLMs to accurately generate fine-grained concept descriptions given an image. To summarize, our contributions include:
\begin{itemize}
    \item We highlight the lack of fine-grained image comprehension ability of instruction-tuned LVLMs across various real-life objects. To the best of our knowledge, we are the first to explore FGVC as an evaluation criteria for these models and their lack of ability thereof.
    \item We underscore the persistence of modality gap in state-of-the-art LVLMs by conducting an extensive per-modality-based probing, revealing the discrepancy in how the two modalities are processed by these models (\S \ref{sec:modality_gap}).
    \item We construct a novel attribute-centric benchmark for FGVC to open up a new direction for future works to measure LVLMs' modality gap and their granular image understanding capability. Our \textbf{\textsc{Finer}} training mixture and newly proposed prompting technique \textbf{\textsc{AttrSeek}} enable substantially improved zero-shot FGVC performance for GPT-4V (\S \ref{sec:brittleness_vlms}) and LLaVA-1.5 (\S \ref{sec:zero_shot_transferability}).
    \vspace{-0.1cm}
\end{itemize}

\section{Related Work}
\subsection{Instruction-tuned Large Vision-Language Models}
State-of-the-art LVLMs such as LLaVA \citep{liu2023improvedllava, liu2023llava}, BLIP-2 \cite{li2023blip}, InstructBLIP \citep{dai2023instructblip}, and closed-source models like GPT-4V \cite{OpenAI_GPT4_2023, yang2023dawn} have brought to our attention their zero-shot task solving abilities, especially in downstream tasks such as Visual Question Answering (VQA), reasoning and image captioning, all of which require output generation conditioned on extensive knowledge of the real-world. Based on an intricate interplay between their parametric knowledge and image understanding ability, they are able to generate sensible outputs. However, many of them focus almost exclusively on  image captioning and reasoning, most often disregarding the concept recognition tasks traditional computer vision tasks evaluate on.

\subsection{Fine-grained Visual Categorization}
% Traditional works in this domain - why they should be solved via zero-shot, modality-gap reduced approach?
Previous works approach FGVC with masked image modeling \citep{he2022masked, ryali2023hiera}, concept meta-information injection \citep{diao2022metaformer}, or LLM-generated concept descriptions \citep{menon2023visual}. However, learning of fine-grained visual categories in LVLMs and their ability to elaborate on the fine-grained details of the input image via text generation is yet to be explored. Furthermore, recent works in FGVC have also shown to disregard fine-grained details of images \citep{krojer-etal-2022-image} and work poorly on downstream tasks involving localization \citep{Ranasinghe_2023_ICCV, zhong2022regionclip} and object attributes \citep{yuksekgonul2022}.

\subsection{Modality Gap in Vision-Language Models}
Different from instruction-tuned VLMs like LLaVA and InstructBLIP, contrastively trained VLMs like CLIP \cite{radford2021learning} and GLIP \citep{li2022grounded} rely on directly minimizing the contrastive objective between visual and textual representations. A recent analytical work \citep{liang2022modalitygap} on CLIP-like models reveals that there is a modality gap between the text and visual modalities, which imposes substantial implications on downstream task performances. Our work shows that such \textit{modality gap} also exists in LVLMs like LLaVA and InstructBLIP, despite the noticeable architectural difference between models like CLIP and LVLMs discussed in this work.

\begin{figure*}[ht]
    \centering
    \includegraphics[width=\textwidth]{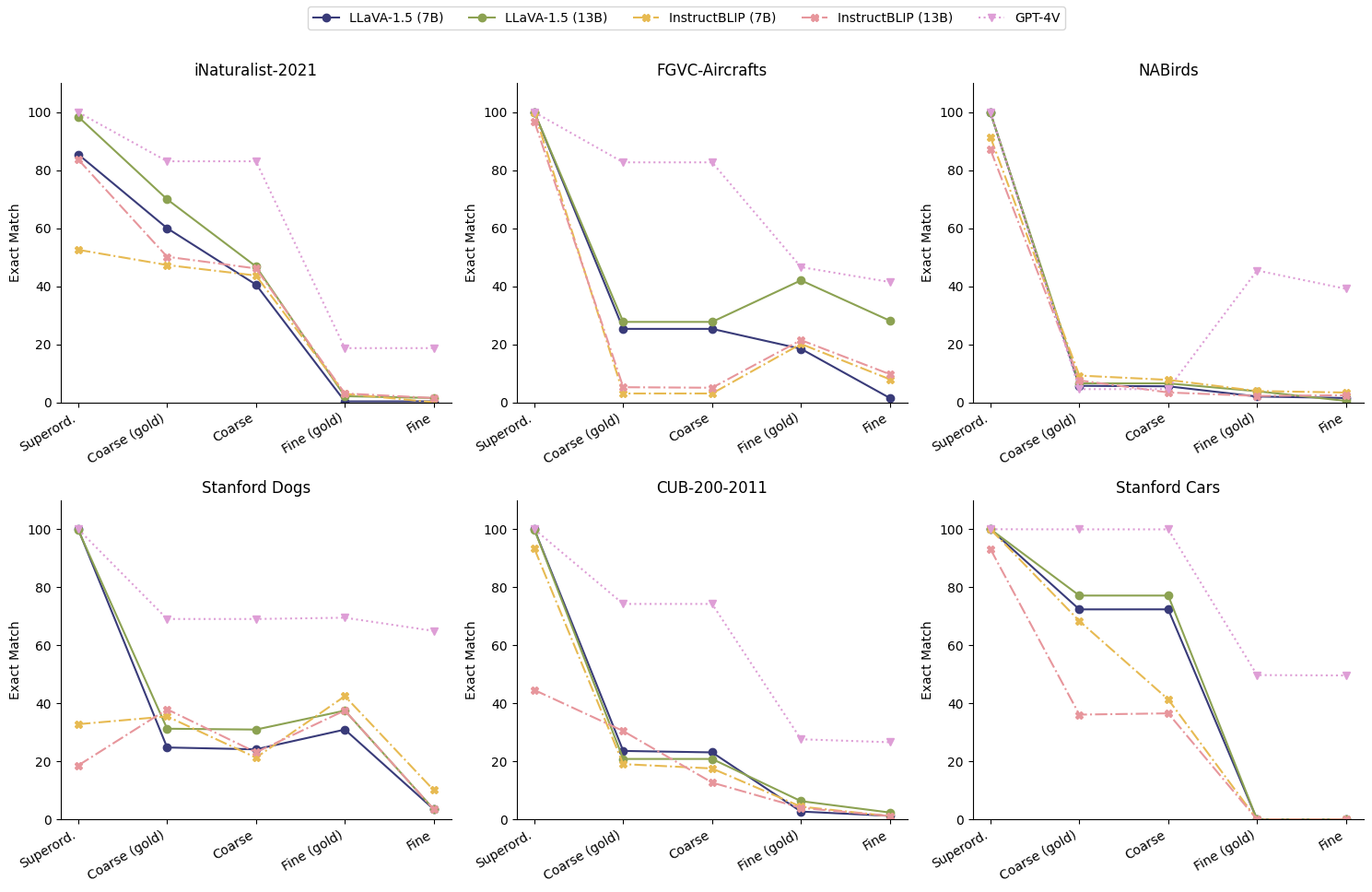}
    \caption{\textbf{State-of-the-art instruction-tuned LVLM zero-shot performance on fine-grained classification.} All the models exhibit strong classification capabilities when prompted to classify superordinate-level (e.g., \textit{birds}, \textit{cars}) and coarse-grained categories(e.g., \textit{owls}, \textit{SUVs}), but exhibit significant deterioration in performance when prompted to categorize more fine-grained categories on the same images. The \texttt{gold} tags for coarse- and fine-grained classifications denote the use of gold labels from the parent category in the prompt.}
    \label{fig:vlm_performance_brittle}
\end{figure*}

\begin{figure}[t]
    \centering
    \includegraphics[scale=0.38]{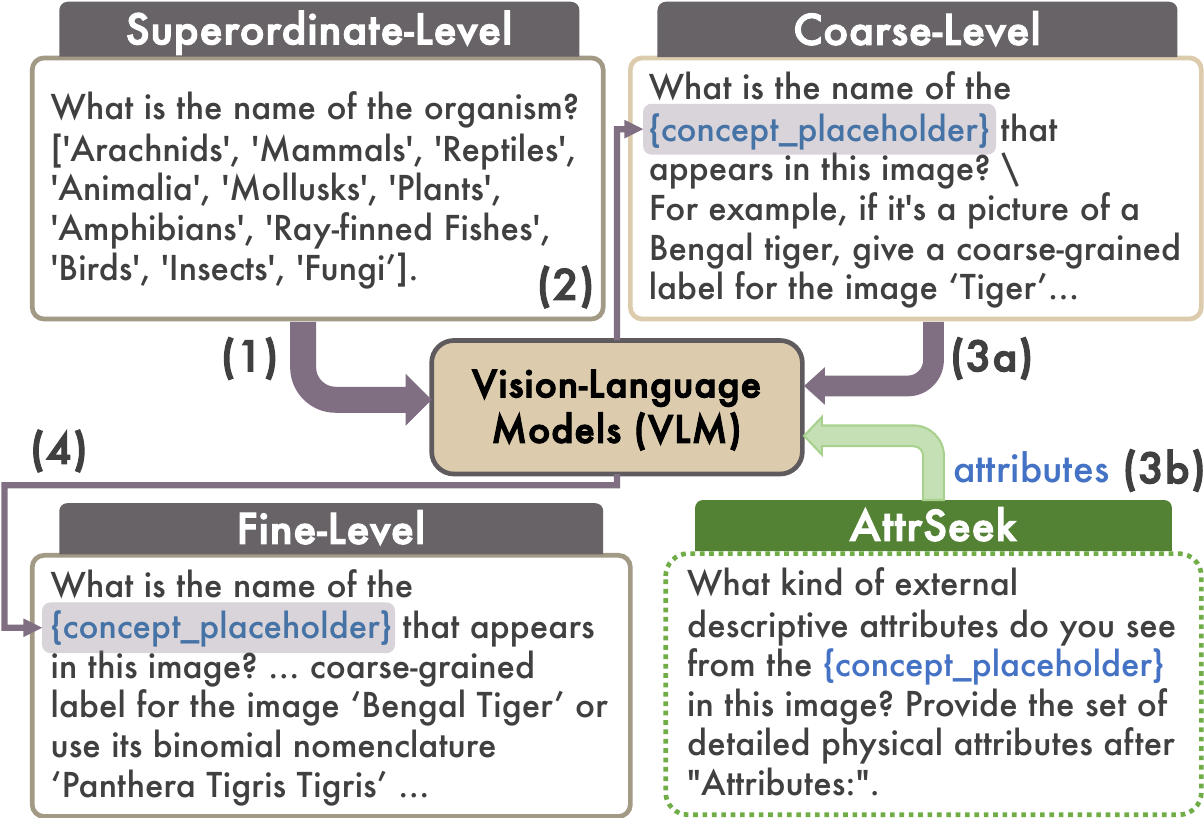}
    \caption{\textbf{Fine-grained classification pipeline.} At each level, an output from LVLM is injected into the next level prompt. \textbf{(1)} Superordinate-level prompt is used to predict the highest-level category (e.g., \textit{bird}). \textbf{(2)} Coarse-level prompt is subsequently fed with the predicted output and fed back to the LVLM to generate the next output (e.g., \textit{parrot}), and \textbf{(3a)} and \textbf{(4)} follow the same steps. \textbf{(3b)} illustrates \textbf{\textsc{AttrSeek}}, a newly proposed prompting scheme in this work, wherein the model is prompted to generate the visual attributes.}
    \label{fig:fgvc_pipeline}
\end{figure}

\section{Fine-Grained Image Understanding in Vision-Language Models}
We first evaluate the fine-grained visual categorization performance of five different instruction-tuned baselines on six different FGVC benchmarks. This section elaborates on the details of the experimental setup and models investigated in this work.

\subsection{Evaluation Settings}
\label{sec:eval_setting}
\paragraph{Datasets} Covering a wide range of real-world objects over various categories, existing FGVC benchmarks provide richly annotated set of image-concept pairs. As shown in Figure \ref{fig:vlm_performance_brittle}, we use iNaturalist-2021 \citep{van2018inaturalist}, FGVC-Aircrafts \citep{maji13fine-grained}, Stanford Dogs \citep{khosla2011novel}, Stanford Cars \citep{krause2013collecting}, NABirds \citep{van2015building}, and CUB-200-2011 \citep{WahCUB_200_2011}. For each dataset, we divide the ground-truth concept label into three levels of granularity: superordinate, coarse and fine, as defined in a previous work \citep{doi:10.1177/0165551513481443}. Superordinate level refers to the highest taxonomic concepts (e.g., \textit{bird}, \textit{car}), coarse level refers to the lower-level granularity concepts (e.g., \textit{parrot}, \textit{SUV}), and fine level refers to the lowest, finer-level granularity (e.g., \textit{owl parrot (Strigops habroptila), Hyundai Santa Fe 2018}). We discuss each benchmark and the construction of superordinate and coarse-grained labels in detail in Section \ref{sec:benchmark_construction}.

\paragraph{Metrics} We assess the accuracy of the generated concept labels given an image and a granularity-specific prompt asking the model to figure out the correct category the concept in the image belongs to. Following previous works on concept classification using auto-regressive models, we employ F1 and Exact Match (EM) scores; note that the EM score used in this work is a modified EM score that parses a sequence of generated text and considers the output label correct if the ground-truth label string exists within a pre-defined maximum number of tokens, $m$ (we set $m=20$).

\paragraph{Models}
The models used in this work are as follows: \textbf{LLaVA-1.5} \citep{liu2023llava, liu2023improvedllava}, \textbf{InstructBLIP} \citep{dai2023instructblip}, and \textbf{\textsc{GPT-4V}}; the hyperparameter setting for each model is in Appendix \ref{sec:appendix_hyperparameters}. The open-sourced models like LLaVA-1.5 and InstructBLIP follow a generic pipeline of transforming an input image $\mathbf{X_{v}}$ with a frozen vision encoder such as CLIP ViT-L/14 \cite{radford2021learning} into an encoded image representation $\mathbf{Z_{v}}$. Then, these models either project $\mathbf{Z_{v}}$ into the language representation space through a learned projection layer $\mathbf{W}$, which becomes $\mathbf{H_{v}} = \mathbf{W} \cdot \mathbf{Z_{v}}$ as in \citet{liu2023llava}, or attend over $\mathbf{Z_{v}}$ with learnable queries $\mathbf{Q}$ as in \citet{li2023blip, dai2023instructblip}. Such transformed visual representations interact with $\mathbf{X_{instruct}}$, a language instruction, which attends over the image representations (or queries) within the self-attention layers of the LLM component to generate the final output sequence.

\subsection{Brittleness of Vision-Language Models}
\label{sec:brittleness_vlms}
\paragraph{Zero-shot Model Performance on FGVC} To evaluate the fine-grained image recognition ability of LVLMs, we measure their classification performance per granularity by prompting the models to generate the correct label for a given concept image as shown in Figure \ref{fig:vlm_performance_brittle}. As illustrated in Figure \ref{fig:fgvc_pipeline}, we assess the models' classification ability across three different granularity levels. In Figure \ref{fig:vlm_performance_brittle}, we 
evidence significant deterioration in terms of classification performance across all the five baselines, with some, e.g., iNaturalist-2021, even reaching near 0\% in EM score. While models do perform very well for superordinate-level categories, often achieving 100\% in EM, the finer granularity leads to substantially worsened classification performance. In terms of model size, larger models like LLaVA-1.5 (13B) and GPT-4V tend to perform better than smaller models like the 7B variants. For InstructBLIP, the 7B version performs better than the 13B version by a large margin, with the 13B version exhibiting less capable instruction-following ability than the 7B one, potentially due to 7B variant being less prone to overfitting and exhibiting efficiency in simple tasks like classification. 

\setlength{\tabcolsep}{2.5pt}
\begin{table}[t!]
\small
\centering
\begin{tabular}{lccc}
\toprule
\textbf{Models} & \textbf{Superordinate} & \textbf{Coarse} & \textbf{Fine} \\
\midrule
\textbf{LLaVA-1.5 (13B)} & 96.872 & \textbf{40.625} & 1.562 \\
\quad+ CoT (0-shot) & - & 35.910 & \textbf{4.687} \\
\quad+ CoT (3-shot) & - & 34.925 & 2.159 \\
\quad+ \textbf{\textsc{AttrSeek}} & - & 35.937 & 1.562 \\
\midrule
\textbf{InstructBLIP (13B)} & 98.437 & \textbf{50.221} & \textbf{3.714} \\
\quad+ CoT (0-shot) & - & \textbf{31.805} & 1.238 \\
\quad+ CoT (3-shot) & - & 30.591 & 0.919 \\
\quad+ \textbf{\textsc{AttrSeek}} & - & 30.177 & 3.020 \\
\midrule
\textbf{GPT-4V} & 100.00 & 83.115 & 18.752 \\
\quad+ CoT (0-shot) & - & 93.750 & 20.312 \\
\quad+ CoT (3-shot) & - & 95.283 & 24.389 \\
\quad+ \textbf{\textsc{AttrSeek}} & - & \textbf{95.331} & \textbf{53.125} \\
\bottomrule
\end{tabular}
\caption{\textbf{Elicitive prompting results (EM;\%) on iNaturalist.} Prompting techniques like Chain-of-Thought (CoT) on the fine-grained classification (\textbf{Fine}) is unable to improve performance in open-source models.}
\label{table:elicitive_prompting}
\vspace{-0.4cm}
\end{table}

\paragraph{Elicitive Prompting for FGVC} LLMs like Vicuna \citep{vicuna2023} and LLama \citep{touvron2023llama} used as textual reasoning components of LVLMs are known to perform better when presented with elicitive prompts like Chain-of-Thought (CoT) \citep{wei2022chain} that improve model's reasoning ability. A unifying thought along this line of prompting techniques is to break down a complex problem into a sequence of sub-problems, i.e., divide-and-conquer. Inspired by these prompting techniques, we propose and evaluate our prompting technique for FGVC, \textbf{\textsc{AttrSeek}}. In this simple prompting strategy, we first (i) prompt the models to generate the most distinctive physical attributes visible in the concepts in an image, and (ii) feed the generated set of attributes along with the concept-asking prompt for final prediction as shown in Figure \ref{fig:fgvc_pipeline}. We evaluate the 13B model variants and GPT-4V only on iNaturalist-2021 dataset due to budget constraint of evaluating on the full evaluation set. In Table \ref{table:elicitive_prompting}, we do not see much improvement in terms of fine-grained concept classification in the open-source models, with neither CoT nor \textsc{AttrSeek} enhancing the classification performance. However, there is a substantial increase in fine-level performance for GPT-4V when prompted with our simple yet effective \textsc{AttrSeek} scheme. This result suggests that LLaVA-1.5 and InstructBLIP, while they exhibit strong image captioning and reasoning ability, are limited in terms of image-grounded attribute understanding even when provided with elicitive prompts; we further elaborate on the need to finetune the model according to the newly proposed prompting scheme in Section \ref{sec:zero_shot_transferability}. This result also suggests that open-source LVLMs may lag behind in instruction-following abilities in comparison to GPT-4V. For additional details on few-shot prompting, see Appendix \ref{sec:appendix_multiple_image_inputs_lvlms}.

\section{Modality Gap: Discrepancy Between Textual and Visual Modalities}
\label{sec:modality_gap}
We hypothesize that the lack of zero-shot concept classification ability of LVLMs arises mainly due to the modality gap between textual and visual inputs, preventing the models from leveraging the existing concept-related parametric knowledge when an image of a concept is given. Note that modality gap studied in this work is different from the one previously identified in CLIP-like VLMs \citep{liang2022modalitygap}.
In this section, we aim to delve into the details of how these models process visual and textual modalities by exploring how well they perform when given only textual descriptions of a concept (\S \ref{sec:probing_concept_knowledge}) and how they accurately elaborate what they see in a given image (\S \ref{sec:attribute_generation_modality_gap}). We also perform linear probing (\S \ref{sec:linear_probing}) against projected and original vision encoder output embeddings to gauge the influence of projection and the subsequent loss of visual information on the modality gap.

\begin{table*}[ht!]
\small
\centering
\begin{adjustbox}{width=\textwidth}
\begin{tabular}{lcccccccc}
\toprule
\multirow{2.5}{*}{\textbf{Base Model}} & \multirow{2.5}{*}{\textbf{LLM}} & \multirow{2.5}{*}{\textbf{Input Modality}} & \multicolumn{5}{c}{\textbf{Similarity Measurement Metrics}} \\
\cmidrule(lr){4-8}
 & & & \textbf{ROUGE-1} & \textbf{ROUGE-2} & \textbf{ROUGE-L} & \textbf{BertScore} & \textbf{AlignScore} \\ 
\midrule

\multirow{5}{*}{\shortstack{\textbf{LLaVA-1.5}\\\citep{liu2023improvedllava}}} 
 & \multirow{2}{*}{\texttt{Vicuna-7B}} & \texttt{Text} & \textbf{22.343} & \textbf{11.023} & \textbf{21.161} & \textbf{85.429} & \textbf{18.935} \\
 & & \texttt{Image} & 17.468 & 10.049 & 17.314  & 84.499 & 15.509 \\
 & \multirow{2}{*}{\texttt{Vicuna-13B}} & \texttt{Text} & \textbf{20.616} & 10.902 & \textbf{19.647}& \textbf{85.944} & \textbf{15.746} \\
 & & \texttt{Image} & 19.661 & \textbf{11.334} & 19.114 & 85.704 & 13.319 \\
 & & $\ \ \Delta$ \texttt{Avg.} & \textcolor{blue}{2.920} & \textcolor{blue}{0.703} & \textcolor{blue}{2.189} & \textcolor{blue}{0.585} & \textcolor{blue}{2.926} \\
 \midrule

 \multirow{5}{*}{\shortstack{\textbf{InstructBLIP}\\\citep{dai2023instructblip}}} 
 & \multirow{2}{*}{\texttt{Vicuna-7B}} & \texttt{Text} & \textbf{18.987} & 9.192 & \textbf{16.577} & \textbf{83.102} & \textbf{15.002} \\
 & & \texttt{Image} & 16.018 & \textbf{9.788} & 11.258 & 82.681 & 10.186 \\
 & \multirow{2}{*}{\texttt{Vicuna-13B}} & \texttt{Text} & \textbf{13.731} & \textbf{7.541} & \textbf{16.446} & \textbf{80.549} & \textbf{13.884} \\
 & & \texttt{Image} & 10.949 & 6.009 & 9.583 & 79.277 & 10.191 \\
 & & $\ \ \Delta$ \texttt{Avg.} & \textcolor{blue}{2.875} & \textcolor{blue}{1.064} & \textcolor{blue}{6.091} & \textcolor{blue}{0.846} & \textcolor{blue}{4.254} \\
 \midrule

 %  \multirow{5}{*}{\shortstack{\textbf{BLIP-2}\\\citep{li2023blip}}} 
 % & \multirow{2}{*}{\texttt{Flan-T5-XL}} & \texttt{Text} & \textbf{20.722} & \textbf{10.511} & \textbf{20.118} & \textbf{85.859} & \textbf{15.827} \\
 % & & \texttt{Image} & 14.107 & 6.173 & 13.376 & 81.343 & 11.673 \\
 % & \multirow{2}{*}{\texttt{Flan-T5-XXL}} & \texttt{Text} & \textbf{23.174} & 
 % \textbf{12.482} & \textbf{21.891} & \textbf{86.720} & \textbf{19.680} \\
 % & & \texttt{Image} & 18.398 & 9.490 & 16.237 & 84.190 & 15.294 \\
 % & & $\ \ \Delta$ \texttt{Avg.} & \textcolor{blue}{5.695} & \textcolor{blue}{3.664} & \textcolor{blue}{6.197} & \textcolor{blue}{3.522} & \textcolor{blue}{4.270} \\ \midrule

 \multirow{3}{*}{\shortstack{\textbf{GPT-4V}\\\citep{OpenAI_GPT4_2023}}} 
 &  & \texttt{Text} & \textbf{24.320} & \textbf{10.179} & \textbf{22.748} & \textbf{87.457} & \textbf{10.524} \\
 & & \texttt{Image} & 22.675 & 8.812 & 20.477 & 85.495 & 5.067 \\
 & & $\ \ \Delta$ \texttt{Avg.} & \textcolor{blue}{1.644} & \textcolor{blue}{1.367} & \textcolor{blue}{2.271} & \textcolor{blue}{1.961} & \textcolor{blue}{5.456} \\

\bottomrule
\end{tabular}
\end{adjustbox}
\caption{Measuring the modality gap via textual similarity against the Web-extracted concept attributes against the LVLM-generated attributes for the iNaturalist subset in \textbf{\textsc{Finer}}. The discrepancy between the attributes generated from \texttt{Text}-only input and \texttt{Image}-only input indicate that VLMs treat the two modalities of the same concept differently. $\ \ \Delta$ \texttt{Avg.} indicates the average difference between the \texttt{Text} and \texttt{Image} outputs against the reference.}
\label{table:text_similarity_modalitygap}
\vspace{-0.3cm}
\end{table*}

\subsection{Probing for Concept-Related Parametric Knowledge}
\label{sec:probing_concept_knowledge}

With the drastic performance drop in Section \ref{sec:brittleness_vlms}, we first need to verify whether concept-attribute knowledge already exists within the LVLM parameters to make sure that the models have already acquired the knowledge necessary for zero-shot classification.
The concept-attribute knowledge refers to the textual parametric knowledge related to specific concepts, e.g., a concept \textit{Bengal Tiger} has visual attributes \textit{dark brown or black stripes}.

\begin{figure}[t]
    \centering
    \includegraphics[scale=0.5]{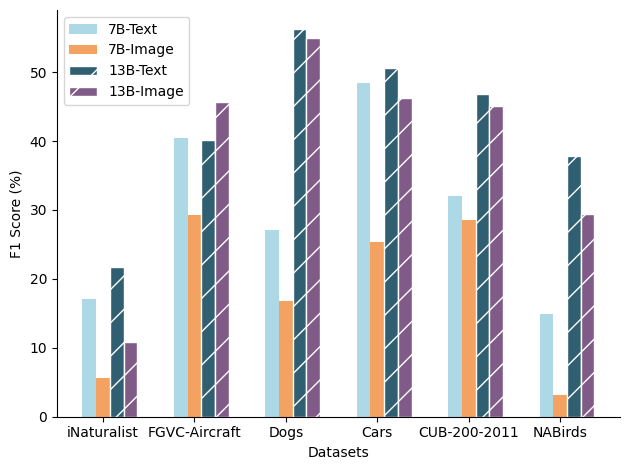}
    \caption{\textbf{Model Performance on Text-only vs. Image-only Inputs.} LLaVA-1.5 (\texttt{7B} and \texttt{13B}), when provided only the textual information (\texttt{7B-}, \texttt{13B-Text}) related to the ground-truth concept, outperforms the image-only input (\texttt{7B-}, \texttt{13B-Image}) counterpart.}
    \label{fig:knowledge_probe_figure}
    \vspace{-0.5cm}
\end{figure}

In this experiment, we measure the classification performance of the models with two different input types: (i) \textbf{Text-only} input that consists of a concept's external visual attributes (details about the attribute extraction in \S \ref{sec:benchmark_construction}), and (ii) \textbf{Image-only} input that consists of the image of the concept. The text-only input, $\mathbf{X_{txt}} = [I; Attr; C]$, is composed of Instruction ($I$), visual attributes ($Attr$), and coarse-grained labels ($C$); the image-only input is  $\mathbf{X_{v}} = [I; X_{img}; C]$. The final output is the concept name. Note, for fair comparison, we also include $C$ as input for image-only probing. Refer to Appendix \ref{sec:appendix_knowledge_probe_prompts} for prompts. In Figure \ref{fig:knowledge_probe_figure}, the results show that even with text-only input that contains the detailed physical attributes of a concept, LVLMs are capable of solving fine-grained visual classification, outperforming the image-only input. The results imply two things: (i) concept-attribute knowledge exists in model parameters, and (ii) while the visual attributes are strongly correlated with the concept, the image modalities are incapable of leveraging the concept-attribute knowledge.
We also see that the larger the model size, the better the performance, relating to the amount of parametric knowledge that resides in the LLMs.

\subsection{Measuring the Modality Gap with Attribute Generation}
\label{sec:attribute_generation_modality_gap}
\begin{figure*}[t!]
    \centering
    \includegraphics[width=0.8\textwidth]{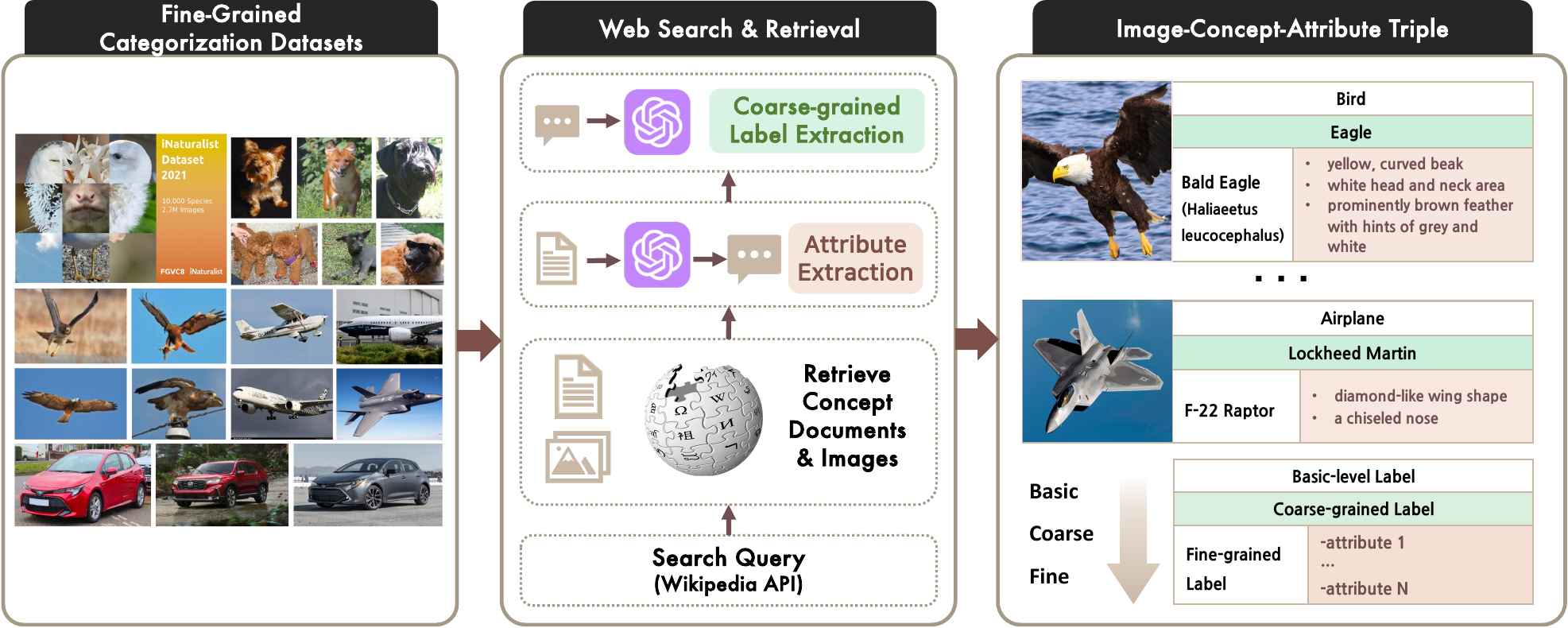}
    \caption{Depiction of the \textbf{\textsc{Finer}} benchmark construction pipeline. Following the aggregation of the six benchmarks in the FGVC domain, concept attributes and concept images are retrieved and extracted from Wikipedia documents. }
    \label{fig:benchmark_pipeline}
\end{figure*}

Having observed the discrepancy between the visual and textual modalities, we now analyze whether LVLMs can observe and tell visually grounded physical attributes of an input image. We construct a set of Web-extracted concept attributes from Wikipedia documents (further elaborated in detail in Figure \ref{fig:benchmark_pipeline}; \S \ref{sec:benchmark_construction}) to be used as the reference texts. Then, we prompt the models to generate a set of external, discriminative physical attributes of a concept when given either an image or text input (see the detailed prompts in Appendix \ref{table:appendix_attribute_generation_prompts}); the text input refers to the concept label along with a prompt that asks for the concept's visual attributes. The textual similarities between the LVLM-generated attributes and the web-extracted attributes are compared using five different scoring metrics that span both the token-level overlap and NLI model-based semantic similarity measure: \textbf{ROUGE-1, 2, L}, \textbf{BertScore} \citep{zhang2019bertscore}, and \textbf{AlignScore} \citep{zha-etal-2023-alignscore}. Both the model-generated attributes and web-extracted attributes are linearized for textual similarity comparison (Appendix \ref{sec:appendix_linearization}). 
%\heng{among these metrics, maybe only BERTscore makes sense. Talk to Steeve and try to add NLI based scores}

As shown in Table \ref{table:text_similarity_modalitygap}, text-only inputs show greater textual similarity to reference attributes, indicating that while the concept-attribute knowledge is being used by the textual modality, the visual modality does not leverage such knowledge to a degree that matches the textual modality. Aside from the modality discrepancy, these models potentially fall short on accurately focusing on specific parts of the image, diluting away the fine-grained details such as stripes or patterns (Table \ref{table:case_examples}) for a coarse-grained understanding of the image.

\subsection{Visual Information Loss After Projection}
\label{sec:linear_probing}
\begin{figure}[t]
    \centering
    \includegraphics[scale=0.5]{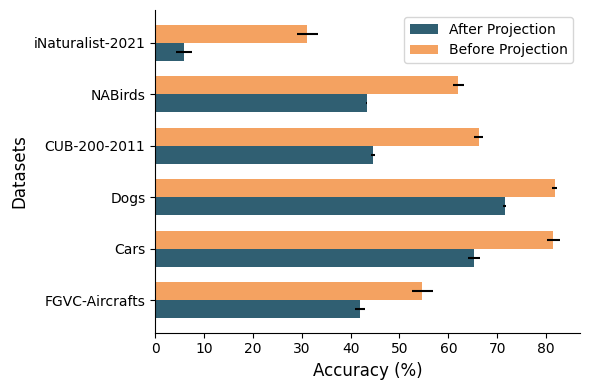}
    \vspace{-0.3cm}
    \caption{\textbf{Linear Probing on Projected Image Embeddings.} Classification accuracy (\%) for before and after image embedding projection to textual space.}
    \label{fig:linear_probing}
    \vspace{-0.5cm}
\end{figure}
We also evaluate the impact of projection from the visual embedding space to the textual space through linear probing. We use CLIP-ViT-L/14 as the image encoder and use LLaVA-1.5's projector. We freeze both the image encoder and the projector and finetune a multi-layer perceptron (MLP) layer on top for classification for 10 epochs (for experiment details refer to Appendix \ref{sec:appendix_linear_probing}). As shown in Figure \ref{fig:linear_probing}, the loss of visual information encoded by the vision encoder leads to substantial drop in classification performance across the six FGVC tasks. The results strongly suggest that such loss of visual information further contributes to the modality gap between the two modalities, especially when performing tasks such as FGVC that rely heavily on fine-grained visual attributes.

\section{\textbf{\textsc{Finer}: Fine-Grained Concept Recognition Benchmark}}
\label{sec:benchmark_construction}
To facilitate research for fine-grained image understanding in LVLMs, we propose a new benchmark and training mixture, \textbf{\textsc{Finer}}. \textbf{\textsc{Finer}} intends to evaluate the interplay between the concept image understanding and attribute knowledge, in addition to the training mixture that mitigates the modality gap and enhances fine-grained concept recognition.

\subsection{Dataset Construction}
We construct \textbf{\textsc{Finer}} based on six different FGVC datasets: iNaturalist-2021 \citep{van2018inaturalist}, FGVC-Aircrafts \citep{maji13fine-grained}, Stanford Dogs \citep{khosla2011novel}, Stanford Cars \citep{krause2013collecting}, NABirds \citep{van2015building}, and CUB-200-2011 \citep{WahCUB_200_2011}. These datasets span a wide range of objects such as airplanes, insects, plants, birds, mammals and cars, challenging the models to cover a variety of fine-grained concepts.

We first crawl all Wikipedia documents and their main images via a search API\footnote{https://github.com/goldsmith/Wikipedia}. We then extract external, visual attributes of a concept (i.e., concept-indicative attributes) with GPT-4V as our attribute extractor \citep{OpenAI_GPT4_2023} for its strong zero-shot text span extraction ability \citep{huang2024critical}. To briefly elaborate, we divide the extracted attributes into two different types: (i) required and (ii) likely attributes. The "required" attributes are the external, concept-indicative attributes that can be used for concept identification, e.g., \textit{blue-tailed hawks with black thorax with a broad apple green stripe}, while the "likely" attributes are attributes that may co-occur with the concept but is not directly correlated with the concept, e.g., \textit{blue-tailed hawks inhabit trees}; we provide the likely attributes as meta-information since existing models such as MetaFormer \citep{diao2022metaformer} has proven that meta-information associated with these fine-grained concepts are beneficial for more accurate FGVC performance; however, since the use of meta-information for better FGVC is not the main focus of this work, we leave it to future works to leverage this information. We also populate the dataset with superordinate and coarse-level concept labels for multi-granular concept recognition performance evaluation. For example, as shown in Figure \ref{fig:benchmark_pipeline}, we assign \textit{Airplane} as the superordinate-level label and \textit{Lockheed Martin} as the coarse-level label since FGVC-Aircrafts dataset provides a concept granularity hierarchy, e.g., Boeing $\rightarrow$ Boeing 707 $\rightarrow$ Boeing 707 MAX. However, datasets like Stanford Dogs do not provide such granular hierarchy for their fine-grained concepts. We therefore few-shot prompt GPT-4V to generate the coarse-level labels and manually inspect their validity. We provide the dataset statistics in Table \ref{table:data_analysis} and extraction prompts in Appendix \ref{sec:appendix_attr_gen_prompts} and \ref{sec:appendix_coarse_lbl_gen_prompts}.

\begin{table*}[ht]
\small
\centering
\begin{adjustbox}{width=\textwidth}
\begin{tabular}{>{\centering\arraybackslash}m{0.30\textwidth} | >{\centering\arraybackslash}m{0.27\textwidth} | >{\centering\arraybackslash}m{0.27\textwidth} | >{\centering\arraybackslash}m{0.27\textwidth}}
\toprule
\textbf{Image (Concept Names)} & \textbf{Attributes} (\texttt{Text}-only) & \textbf{Attributes} (\texttt{Image}-only) & \textbf{\textsc{Finer} Attributes} \\
\toprule
\shortstack[c]{\textbf{Dragonfly | Orthetrum Triangulare}\\ \includegraphics[width=0.25\textwidth]{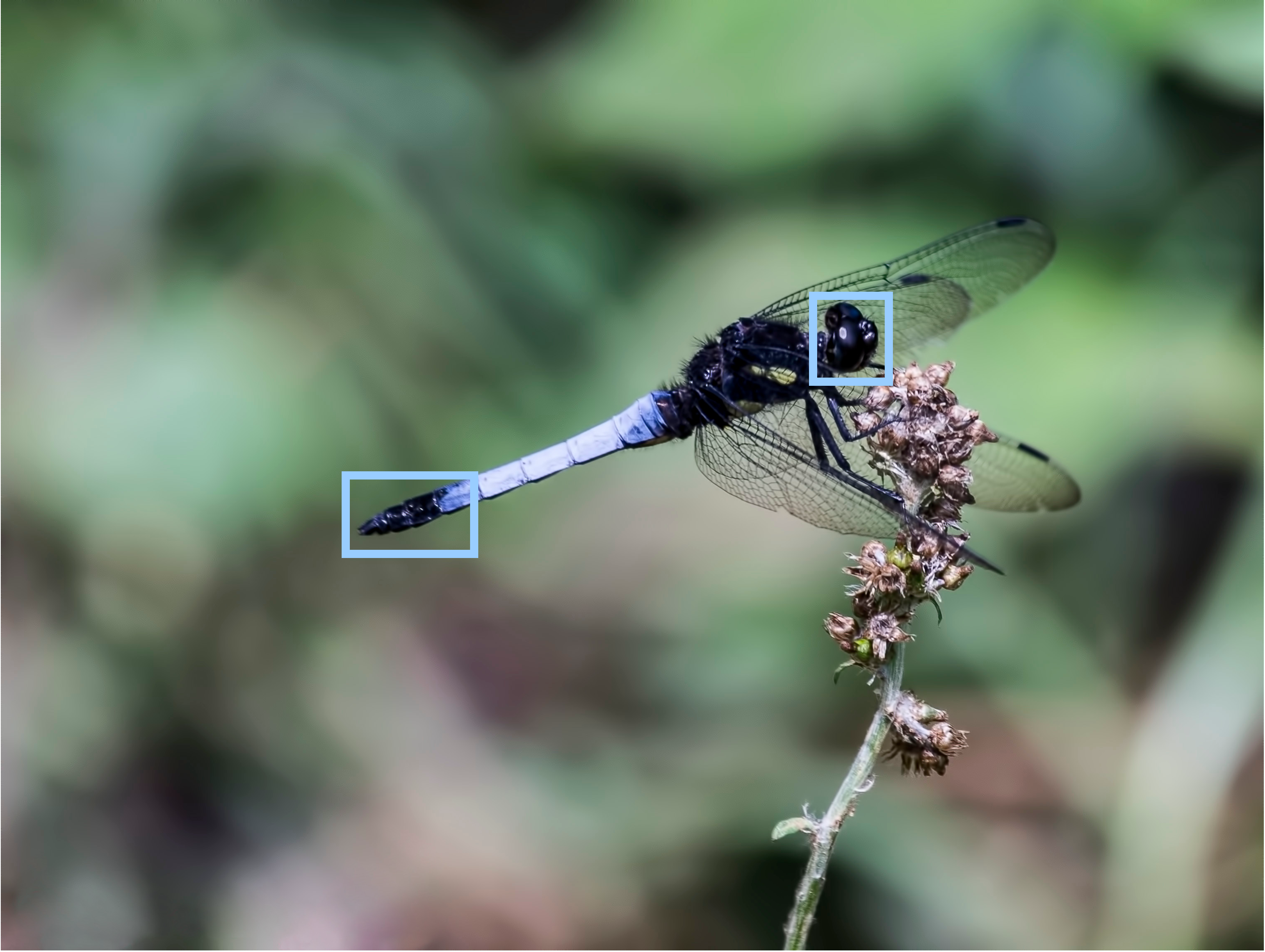}} & \setlength{\fboxsep}{1pt}\colorbox{lightblue}{Blue tail}; Broad wings; White throat;\colorbox{lightblue}{Dark head; Banded tail} &  \setlength{\fboxsep}{1pt}\colorbox{lightred}{Elongated body; Two pairs of} \colorbox{lightred}{wings; Six legs, Compound eyes}, \colorbox{lightred}{Segmented abdomen} & Dark face; Bluish eyes; Black thorax with a broad apple green stripe on both sides; Black segments 1-2 and 8-10 on the abdomen; Remaining segments of the abdomen pruinosed with azure blue \\
\midrule

\shortstack[c]{\textbf{Pinscher | Affenpinscher}\\ \includegraphics[width=0.25\textwidth]{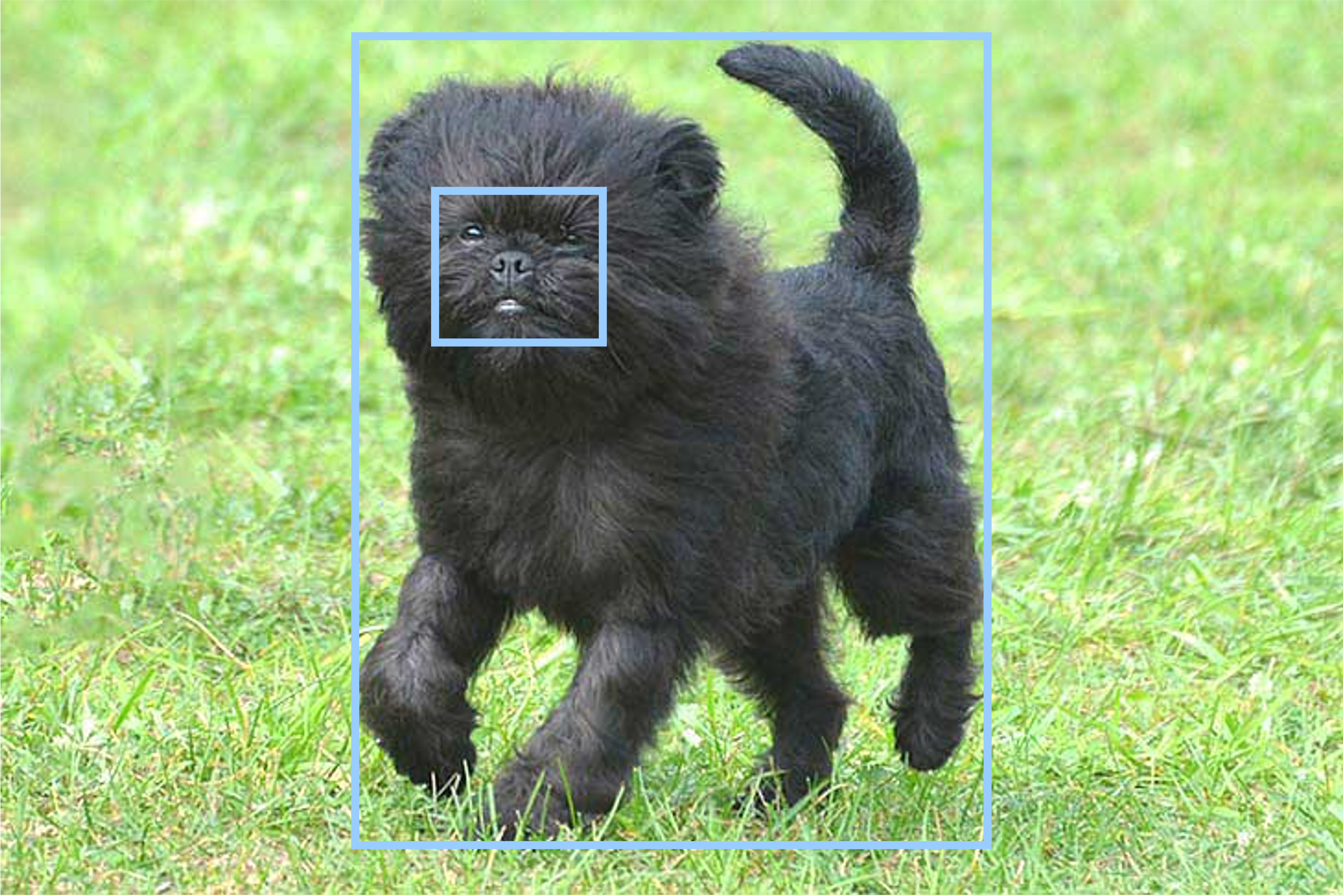}}
 & \setlength{\fboxsep}{1pt}\colorbox{lightblue}{Black}, gray, silver, or tan fur; \colorbox{lightblue}{Rough, shaggy coat; Distinct} \colorbox{lightblue}{"monkey-like" facial expression}; Prominent chin and jaw; Small, round, dark eyes; Ears set high and usually cropped to a point & \setlength{\fboxsep}{1pt}\colorbox{lightred}{Presence of fur; Four-legged} \colorbox{lightred}{stance; Distinct muzzle with a} \colorbox{lightred}{nose; Visible tail}; Ears that are either erect or floppy & Harsh rough coat when not clipped; Shaggier coat over the head and shoulders forming a mane; Shorter coat over the back and hind quarters; Notable monkey-like expression; Coat is harsh and wiry in texture when properly maintained \\
\midrule

\shortstack[c]{\textbf{Embraer | Embraer ERJ 145}\\ 
\includegraphics[width=0.25\textwidth]{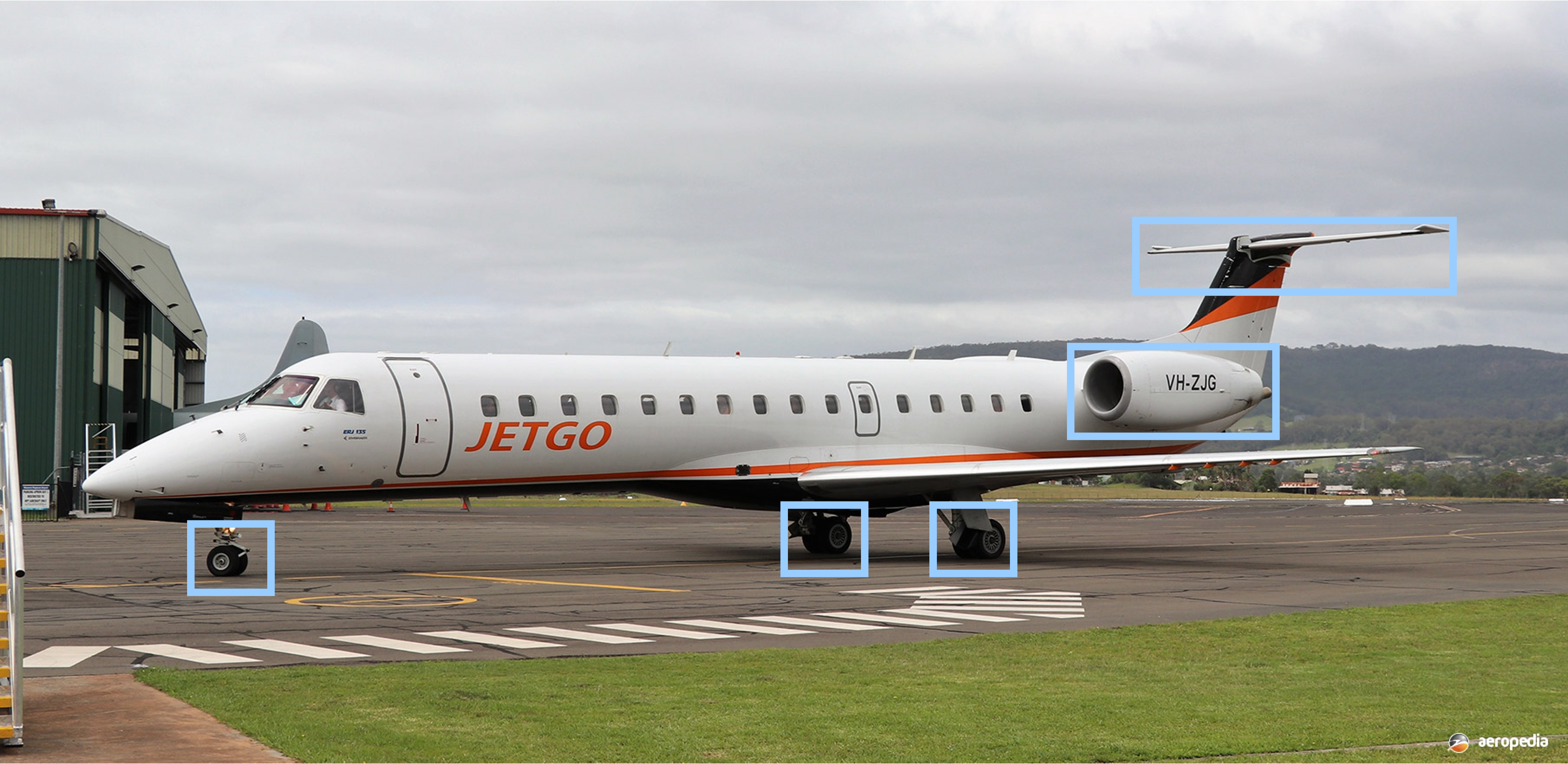}}
 & \setlength{\fboxsep}{1pt}\colorbox{lightblue}{T-tail design}; Straight wing design with no winglets; \colorbox{lightblue}{Mounted engines on the rear} \colorbox{lightblue}{fuselage}; Small cockpit windows compared to the body size; \colorbox{lightblue}{Three sets of landing gear} &  \setlength{\fboxsep}{1pt}\colorbox{lightred}{Fixed wings on either side of the} \colorbox{lightred}{fuselage}; \colorbox{lightred}{Cockpit windows at the front of} \colorbox{lightred}{the fuselage}; \colorbox{lightred}{Jet engines, either} \colorbox{lightred}{under the wings} or mounted on the rear of the fuselage; \colorbox{lightred}{Landing gear} \colorbox{lightred}{with wheels for taking off} & T-tail configuration; Two rear-mounted Rolls-Royce AE 3007 series turbofan engines; Straight wing with no winglets; Narrow, tube-like fuselage; Short, nearly oval-shaped passenger windows.\\
\bottomrule
\end{tabular}
\end{adjustbox}
\vspace{-0.2cm}
\caption{\textbf{Qualitative Analysis of the} \texttt{Text}\textbf{-only and} \texttt{Image}\textbf{-only generated attributes from GPT-4V against \textsc{Finer}.} The generated attributes based on the \texttt{Text}-only and \texttt{Image}-only dataset exhibits notable discrepancy. Juxtaposed with our \textbf{\textsc{Finer} Attributes}, \texttt{Image}-only attributes are not concept-indicative and \setlength{\fboxsep}{1pt}\colorbox{lightred}{generic} compared to \texttt{Text}-only which are \setlength{\fboxsep}{1pt}\colorbox{lightblue}{discriminative} of the concept. We provide \textbf{\textsc{Finer}} attributes as a reference for comparison. We provide the coarse-level and fine-level labels along with the images, and the bounding boxes are only drawn to match the highlighted text and are not part of the dataset.}
\label{table:case_examples}
\end{table*}

\subsection{Qualitative Analysis}
% \heng{since improvement is still small, maybe add some future work about possibly leveraing inter-dependency among attributes and use probabilistic logical reasoning over multiple attributes. You can say that right now you are not using sophisticated reasoning because these rules need to be learned from a lot of annotations. You can take a look at ECOLE proposal after the deadline}

In Table \ref{table:case_examples}, we provide model-generated attributes as a case study on lack of visually-grounded generation. The attributes are from GPT-4V, but note that the generated attributes from other models, such as LLaVA-1.5, all exhibit a similar trend. When provided with image-only input, the model generates a set of attributes that pertain to the image, e.g., elongated body and two pairs of wings for dragonfly. However, the attributes from image-only input are non-discriminative compared to those from text-only input; furthermore, changing the prompting technique to elicit more fine-grained, detailed physical attributes from the LLMs either lead to hallucination or needlessly verbose outputs that describe non-concept related aspects of the input image. In other words, these attributes do not serve as useful knowledge to identify the input image as a specific concept. This again suggests that these models not only fail to properly observe the fine-grained details of a concept, but fail to leverage the knowledge contained within its own parameters as can be seen from the outputs of text-only inputs. 

\subsection{Enhanced Zero-Shot Transferability from Learning to Generate Attributes}
\label{sec:zero_shot_transferability}

\setlength{\tabcolsep}{2.5pt}
\begin{table*}[ht!]
\small
\centering
\begin{tabular}{lcccccc}
\toprule
\textbf{Models} & \textbf{CUB-200-2011} & \textbf{Stanford Dogs} & \textbf{NABirds} & \textbf{iNaturalist} & \textbf{Stanford Cars} & \textbf{FGVC-Aircrafts} \\
\midrule
LLaVA-1.5 (7B) & 5.262 & 16.718 & 3.145 & 6.788 & 25.317 & 29.314\\
Direct Prediction & \textbf{21.071} & 22.942 & 12.610 & 7.109 & 24.624 & 28.622\\
\textbf{\textsc{Finer}} & 20.673 & \textbf{36.297} & \textbf{13.692} & \textbf{7.530} & \textbf{29.974}& \textbf{32.293} \\
\bottomrule
\end{tabular}
\caption{\textbf{Zero-shot Performance on FGVC}. Finetuning on the \textbf{\textsc{Finer}} mixture significantly enhances the zero-shot performance on all six FGVC tasks. We choose LLaVA-1.5 (7B) for this experiment with 1 dataset held-out and 5 other datasets held-in for finetuning. Direct Prediction refers to the setting without the \textbf{\textsc{AttrSeek}} pipeline and the model directly predicting the final concept label without the intermediate attribute generation}
\label{table:zeroshot_transferability_test}
\vspace{-0.1cm}
\end{table*}

To substantiate the effectiveness of our visual attributes in \textbf{\textsc{Finer}}, we construct an instruction-tuning mixture based on the \textsc{AttrSeek} prompting pipeline to improve the zero-shot attribute generation and FGVC performance (see Appendix \ref{sec:appendix_inst_mix_dataset}). The \textbf{\textsc{Finer}} mixture consists of six subsets, where each split is a 5 held-in and 1 held-out FGVC datasets for training and evaluation, respectively. For instance, for the iNaturalist subset, the iNaturalist set is not included in training for zero-shot evaluation. Each instance of the training mixture follows the \textsc{AttrSeek} pipeline (\S \ref{sec:brittleness_vlms}). In Table \ref{table:zeroshot_transferability_test}, we finetune LLaVA-1.5 (7B) on the training mixture and see that the \textbf{\textsc{Finer}}-tuned model outperforms the direct finetuned counterpart that was simply trained to directly predict the concept label. This implies that in FGVC, instruction-tuning LVLMs to attend to visible attributes in images by explicitly generating the attributes and then subsequently performing classification improves the model's performance. Our interpretation is that the generation of the attributes of concepts in the images allow the model to leverage its concept-attribute parametric knowledge (identified in Figure \ref{fig:knowledge_probe_figure}), and perform better zero-shot FGVC classification.
It also underscores the effectiveness of the \textsc{AttrSeek} pipeline in model training to improve the inherent, zero-shot capability of LVLMs for fine-grained concept recognition. We demonstrate case studies in Table \ref{table:case_examples_finetuned}, where we present the zero-shot generated outputs of the \textbf{\textsc{FINER}} mixture-trained LLaVA-1.5 (7B) and the one that was only finetuned to directly predict the final concept label. Refer to Appendix \ref{sec:appendix_experiment_details} for additional details on training.

\begin{table*}[t!]
\small
\centering
\begin{adjustbox}{width=\textwidth}
\begin{tabular}{>{\centering\arraybackslash}m{0.30\textwidth} | >{\centering\arraybackslash}m{0.27\textwidth} | >{\centering\arraybackslash}m{0.27\textwidth} }
\toprule
\textbf{Image (Concept Names)} & 
\textbf{Attributes from \textsc{Finer}} & \textbf{Attributes from Direct Pred.}  \\
\toprule
\shortstack[c]{\textbf{Warbler | Yellow-headed Warbler}\\ \includegraphics[width=0.17\textwidth]{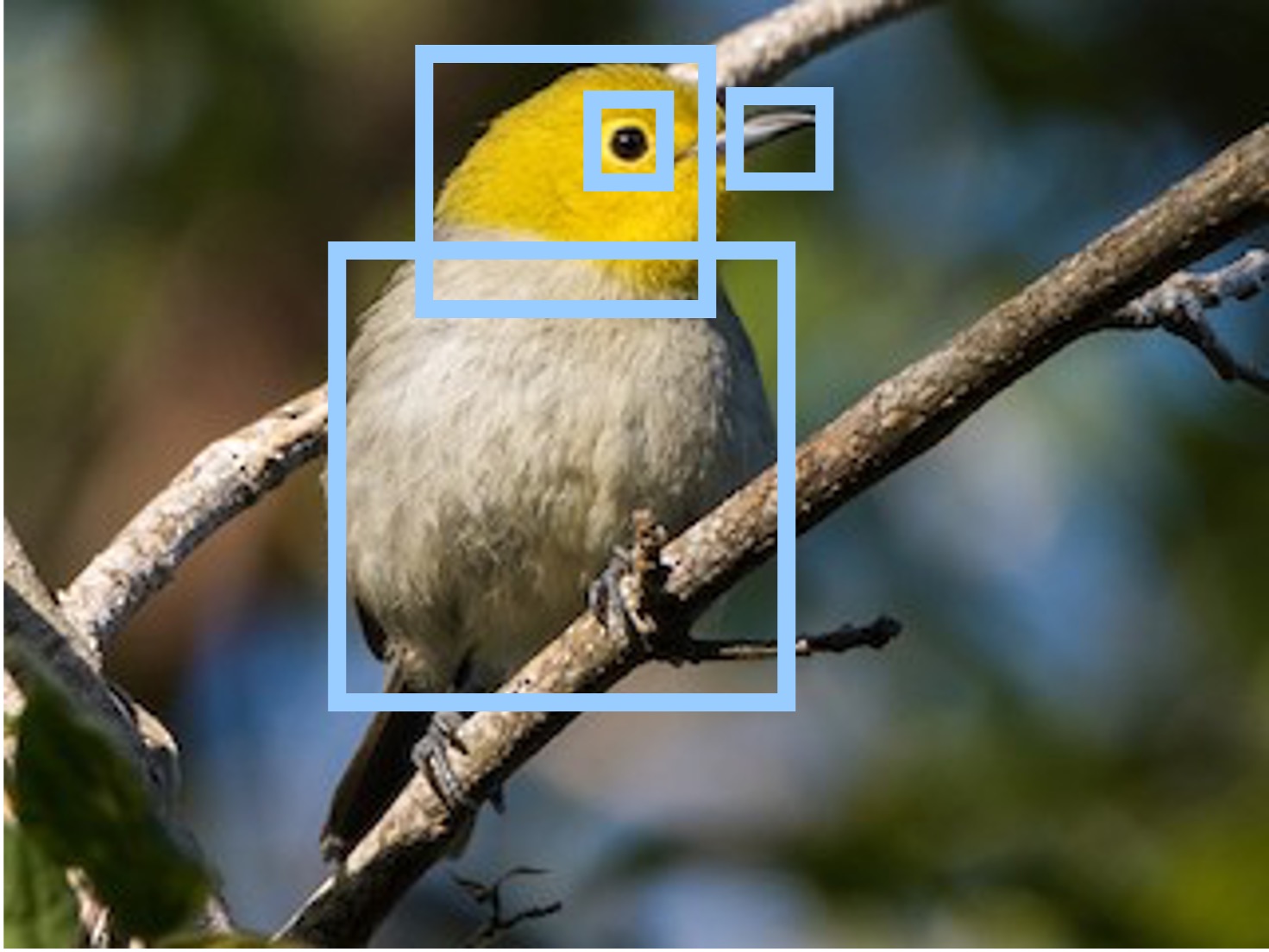}} & 
\setlength{\fboxsep}{1pt} It has \colorbox{lightblue}{yellow round head, white body} with a hint of gray, black eyes and black-colored tail. & \setlength{\fboxsep}{1pt} In the image, the bird is perched on a wire, and it has a \colorbox{lightred}{yellow beak}. Its physical characteristics include a white body, a black tail, and a \colorbox{lightred}{yellow beak}. \\
\midrule

\shortstack[c]{\textbf{Albatross | Black-footed Albatross}\\ \includegraphics[width=0.15\textwidth]{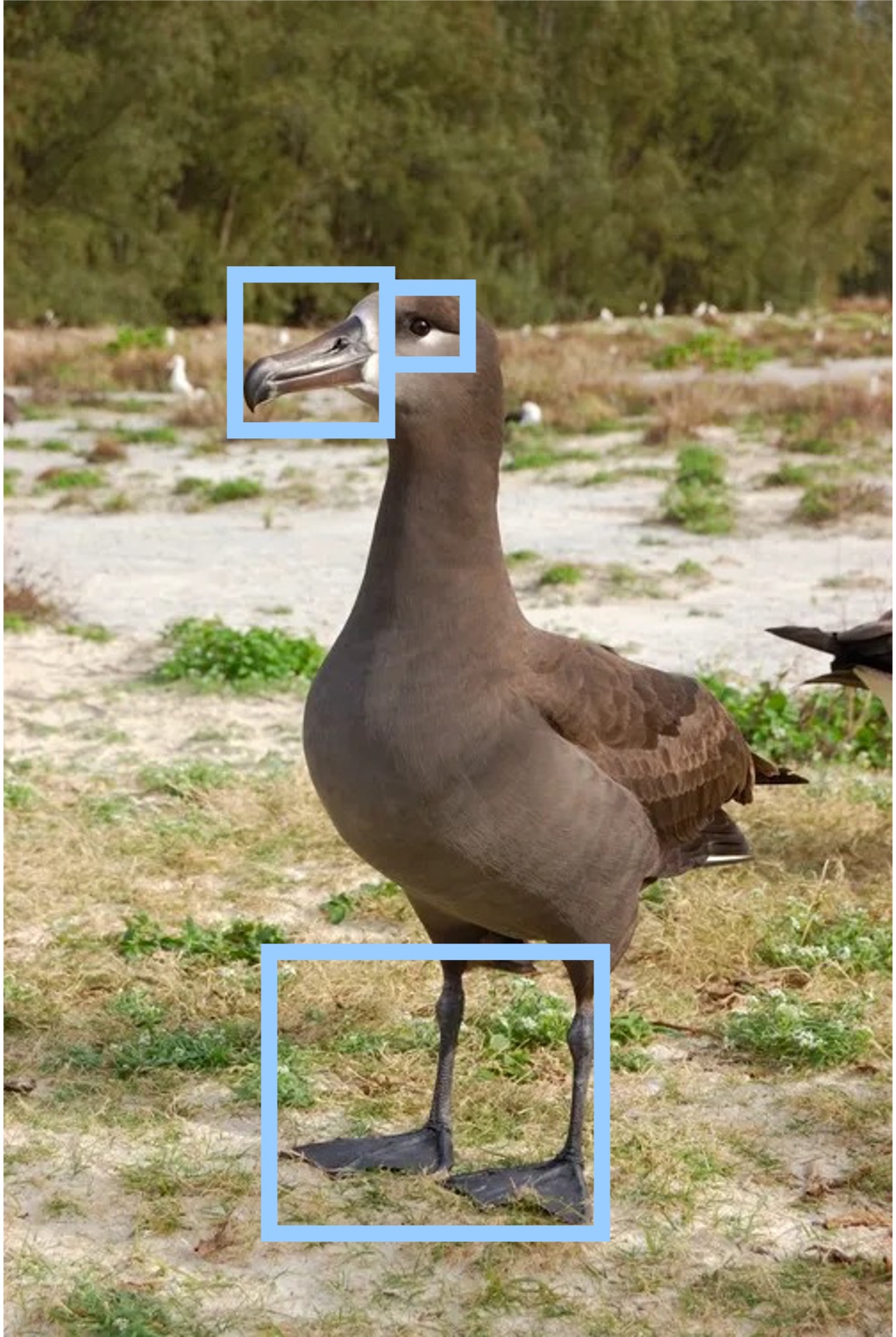}}
 & \setlength{\fboxsep}{1pt}\colorbox{lightblue}{Almost all black plumage, White} \colorbox{lightblue}{markings around the base of the} \colorbox{lightblue}{beak, White markings below the} \colorbox{lightblue}{eye, Dark beak and feet}, White undertail coverts in some adults. &  \setlength{\fboxsep}{1pt} The bird has \colorbox{lightblue}{Predominantly dark color with} subtle variations between \colorbox{lightred}{brown and gray}. The face appears to have lighter shades of \colorbox{lightred}{brown or gray}. \\

\bottomrule
\end{tabular}
\end{adjustbox}
\vspace{-0.1cm}
\caption{\textbf{Qualitative analysis of the zero-shot outputs generated by LLaVA-1.5 (7B) instruction-tuned on \textsc{Finer} and Direct Prediction dataset.} The generated attributes from the \textbf{\textsc{Finer}} trained LLaVA-1.5 (7B) generates accurate, image-grounded outputs when compared to the Direct Prediction counterpart. The attributes are indicated by their \setlength{\fboxsep}{1pt}\colorbox{lightblue}{discriminative} characteristics in contrast to more \setlength{\fboxsep}{1pt}\colorbox{lightred}{generic/hallucinated} ones. We provide the coarse-level and fine-level labels along with the images, and the bounding boxes are only drawn to match the highlighted text and are not part of the dataset.}
\label{table:case_examples_finetuned}
\end{table*}

\section{Discussion and Conclusion}
In this paper, we provide an in-depth analysis on the substantial lack of fine-grained visual comprehension (FGVC) ability among instruction-tuned LVLMs. Our work discovers the presence of modality gap in LVLMs 
and propose a prompting scheme, \textsc{AttrSeek}, and training mixture, \textbf{\textsc{Finer}} to mitigate the gap and improve the zero-shot FGVC ability of LVLMs. We also reveal that the loss of visual information after projection hinder the effective cross-modal interplay from manifesting. Such discrepancy leads to LVLMs being unable to exploit the rich parametric knowledge and deteriorates performance in visual concept recognition. While this is the first study of FGVC among instruction-tuned LVLMs, we hope our work would further the research endeavors in this direction.

%%%%%%%%%%%%%%%%%%%% Limitations %%%%%%%%%%%%%%%%%%%%%%%

\section{Limitations}
\paragraph{Intra-Concept Variance in Images} Images of a single concept can appear in various different forms. Some images may have the whole view of the concept, while other images may have certain parts of the concept (e.g., legs, wings) partially occluded. The attributes collected per concept in \textbf{\textsc{Finer}} are constructed to be visually-grounded based on the textual attributes extracted from Web documents that pertain to these concepts; nonetheless, such intra-concept variance among images may render an attribute obsolete for certain images. Other problems such as low-quality images may also lead to this issue. In future work, exploring the visual "ground-ability" of each attribute through image-text retrieval may be a plausible approach to identifying both the most discriminative attributes that pertains to an image and the fine-grained concept label.

\paragraph{Selection of Baseline Models} While our work covers LVLMs that receive image and text as input, there are other VLMs such as Kosmos-2 \citep{peng2023kosmos}, Shikra \citep{chen2023shikra} and Ferret \citep{you2023ferret} out there, that receive bounding-box annotated images as input. However, our work only deals with un-marked image and prompt inputs, since the objective of this research is to see whether LVLMs without any referring markings (e.g., bounding boxes), can ground their generative capabilities on the input image.

\section*{Acknowledgement}
We thank the anonymous reviewers for their suggestions and comments. We also would like to thank Ansel Blume, Derek Hoiem and Carl Vondrick for their ideas, feedback and support for this work.
This research is based upon work supported by U.S. DARPA ECOLE Program No. \#HR00112390060. The views and conclusions contained herein are those of the authors and should not be interpreted as necessarily representing the official policies, either expressed or implied, of DARPA, or the U.S. Government. The U.S. Government is authorized to reproduce and distribute reprints for governmental purposes notwithstanding any copyright annotation therein.

\bibliography{anthology,custom}
\bibliographystyle{acl_natbib}

\clearpage
\appendix

\section{Experiment Details}
\label{sec:appendix_experiment_details}
In this section, we elaborate in detail the experimental settings of our work, including the hyperparameter settings of the large vision-language models (LVLMs), and the large language models (LLMs), which are used as a major driving block of these LVLMs. In addition to the experiment settings, we also provide details on the training mixture used in \ref{sec:zero_shot_transferability} and the fine-tuning settings of LLaVA-1.5 (7B).

\subsection{Hyperparameter Settings of Vision-Language Models}
\label{sec:appendix_hyperparameters}
\paragraph{Inference Settings} Our work deals with the inference-time fine-grained image understanding abilities of VLMs. The hyperparameter settings provided in the table are only for inference time (\S \ref{sec:brittleness_vlms}, \ref{sec:modality_gap}, \ref{sec:probing_concept_knowledge}) and are not to be used for fine-tuning.

\setlength{\tabcolsep}{3pt}
\begin{table}[h!]
\small
\begin{center}
\begin{tabular}{lcccc}
    \toprule
    \textbf{Hyperparameter} & \textbf{LLaVA}   &
    \textbf{InstBLIP}  & \textbf{BLIP-2}  &
    \textbf{GPT-4}  \\
    \midrule
    max\_seq\_len   & 256 & 256 & 256 & 256 \\
    top\_p & 1.0 & 0.95 & 0.95 & 1.0 \\
    temperature  & 0.2 & 0.75 & 0.75 & 0.2 \\
    \bottomrule
\end{tabular}
\end{center}
\caption{Hyperparameters of the Vision-Language Models (VLMs) studied in this work. \textbf{LLaVA} refers to LLaVA-1.5 and \textbf{InstBLIP} refers to the InstructBLIP model. All the VLMs use the same hyperparameter settings for both the 7B and 13B variants; the same applies to \texttt{Flan-T5-XL} and \texttt{Flan-T5-XXL} models, which consist of 3B and 11B parameter, respectively.}
\label{table:appendix_hyperparameter}
\end{table}

For LLaVA-1.5 models, we use \texttt{Vicuna-v1.5}, and for InstructBLIP models, we use \texttt{Vicuna-v1.1}; these are the original settings of both of these models.

\paragraph{Fine-Tuning Settings} In Table \ref{table:appendix_hyperparameter_training}, we also provide the hyperparameter settings of the instruction-tuned LLaVA-1.5 (7B) in Section \ref{sec:zero_shot_transferability}. The fine-tuning was conducted with LoRA \citep{hu2021lora} on 4 V100 (16G) for approximately 28 hours on the training mixture, which is elaborated in detail in Section \ref{sec:appendix_inst_mix_dataset}.

\setlength{\tabcolsep}{3pt}
\begin{table}[h!]
\small
\begin{center}
\begin{tabular}{lc}
    \toprule
    \textbf{Hyperparameter} & \textbf{LLaVA-1.5 (7B)}  \\
    \midrule
    max\_seq\_len  & 256 \\
    top\_p & 1.0 \\
    temperature & 0.2 \\
    lora\_r & 128 \\
    lora\_alpha & 256 \\
    gradient\_accum. & 16 \\
    batch\_size & 1 \\
    learning\_rate & 2e-4 \\
    lr\_schedule & cosine decay \\
    optimizer & AdamW \\
    weight\_decay & 0.0 \\
    warmup\_ratio & 0.03 \\
    bf16 & \texttt{False} \\
    fp16 & \texttt{True} \\
    \bottomrule
\end{tabular}
\end{center}
\caption{Hyperparameters of LLaVA-1.5 (7B) for Instruction-tuning on the \textbf{\textsc{Finer}} training mixture.}
\label{table:appendix_hyperparameter_training}
\end{table}

\subsection{Additional Details on Modality Gap Experiment}
\paragraph{Using LLMs for Text-Only Setting}
In Section \ref{sec:modality_gap}, we experiment on the LVLMs' ability to generate textual attributes based on either the text-only input or the image-only input. The caveat of using InstructBLIP and BLIP-2 models is that this models do not simply allow text-only input like LLaVA-1.5 or GPT-4V. We therefore opt to use their LLM components, Vicuna-7B and -13B, Flan-T5-XL and -XXL to generate the attributes for the text-only input. The validity of this setting holds since our goal is to compare the modality gap via how much the model stores the concept-related knowledge within its parameters.
\paragraph{Metrics} For ROUGE, we used the Python's ROUGE API\footnote{https://pypi.org/project/rouge/} for calculation, and we only use the F1 score in this work. As for the model-based metrics, for BertScore \citep{zhang2019bertscore} we use the \texttt{bert-base-uncased} and for AlignScore \citep{zha-etal-2023-alignscore} we use \texttt{roberta-large}. These model-based metrics are state-of-the-art models for textual faithfulness evaluation, making them fit to evaluate both the faithfulness and textual consistency of the generated text.

\paragraph{Linearization of the Web-Extracted Attributes}
\label{sec:appendix_linearization}
In this section, we elaborate on the linearization process of the Web-extracted attributes, which are used as ground-truth reference texts in Section \ref{sec:modality_gap}. The linearization of the attributes is a simple process of concatenating the fine-grained concept name and the generated attributes of the concept. For example, for a dragonfly named \textit{Orthetrum Triangulare}, we construct a linearized string that says, \textit{Orthetrum Triangulare exhibits blue eyes, black thorax with a broad apple green stripe}. The format is as follows: [\texttt{<concept-name>}; \texttt{exhibits}; \texttt{attribute-1}; \texttt{attribute-2}; \texttt{...}; \texttt{attribute-j}]; the list is converted into a single sentence for evaluation;. The same process applies to both the LVLM-generated attributes and Web-extracted attributes to compare their textual similarity in Section \ref{sec:modality_gap}.

\subsection{Enabling Multiple Image Inputs in LVLMs for Few-Shot Prompting}
\label{sec:appendix_multiple_image_inputs_lvlms}
The three open-sourced LVLMs investigated in our work, LLaVA-1.5 \citep{liu2023llava, liu2023improvedllava}, InstructBLIP \citep{dai2023instructblip} and BLIP-2 \citep{li2023blip} were all pre-trained and instruction-tuned based on a single image input and a consecutive sequence of pertinent textual instructions. Nonetheless, these models are capable of receiving multiple images given its input format. For LLaVA-1.5, we simply provide the few-shot ($k$) examples sampled from iNaturalist dataset in by interleaving $k$ \texttt{<image>} tokens along with the input prompt and their ground-truth concept labels. For InstructBLIP and BLIP-2, since they require attending over the input image via cross-attention with a pre-defined set of query embeddings, we first embed each of the $k$ few-shot sample images with the vision encoder. Then, the image embeddings are passed through the Q-Former \citep{dai2023instructblip, li2023blip} to generate an instruction-attended query embeddings that contains the image information, and we concatenate them together to use them as few-shot samples.

\subsection{Construction of the Instruction-Tuning Mixture}
\label{sec:appendix_inst_mix_dataset}
To evaluate the validity of our proposed benchmark, \textbf{\textsc{Finer}}, we construct six different instruction-tuning mixtures on top of LLaVA-1.5's instruction-tuning mixture. For example, to build a training mixture to train the model for iNaturalist evaluation, i.e., the \textbf{\textsc{Finer}}'s iNaturalist subset, we hold out the iNaturalist dataset for evaluation and include the rest of the five other FGVC datasetes into the instruction-tuning mixture. We use all the six FGVC datasets to construct the mixture, sampling 2.5K instances from each of the datasets and their attributes into the training data; note, the 2.5k instances are sampled to follow a uniform distribution for the number of classes for each dataset. We structure each instruction-tuning instance into the \textsc{AttrSeek} pipeline format, with each instruction consisting of three turns: (i) Asking the model for the coarse-level concept category given the superordinate concept, \textit{``Can you identify the bird shown in this image?"}; (ii) asking the model to generate a set of external, descriptive visual attributes of the coarse-level concept, \textit{``What kind of external descriptive attributes do you see from the penguin"}, and finally (iii) predicting the fine-grained concept category given the coarse-level concept and the self-generated attributes set. For each of the three steps, we use GPT-4V to generate 15 possible paraphrases of the instruction in order to avoid biasing the model to specific textual instructions and to retain the model's instruction-following ability. We trained the models for 1 epoch each; the checkpoint for 1 epoch is the one we used to evaluate the FGVC performance in Table \ref{table:zeroshot_transferability_test}.

\section{Prompts}
In this section, we provide the input prompts used in each of the experiments, including the fine-grained visual classification and the elicitive prompting of LVLMs in Section \ref{sec:brittleness_vlms}, probing for concept-related parametric knowledge \ref{sec:probing_concept_knowledge}, attribute generation \ref{sec:attribute_generation_modality_gap}

\subsection{Fine-Grained Visual Classification Prompts}
\label{sec:appendix_fine_grained_prompts}
We structure our prompts as shown in Table \ref{table:appendix_fgvc_prompt}. For datasets with less than 100 class categories, we provide them along with the instruction, allowing the models to choose from the provided list of classes. Therefore, for superordinate-levels and certain coarse-levels, we provide the categories as lists so that the models solve the class generation problem by choosing from the input prompt; this is analogous to a multiple choice setting. However, for fine-grained classes, it is difficult to feed in all the concept categories in the input prompt, since some of the datasets like the iNaturalist-2021 has 10,000 categories to choose from. In order to confine the generation scope for the fine-grained labels, we decided to input the coarse-level label as denoted in Table \ref{table:appendix_fgvc_prompt}, to condition the generation of the fine-grained output within a specified category space.

\begin{table*}[h!]
\begin{center}
\begin{adjustbox}{width=1.0\textwidth}
\begin{tabular}{m{12cm}}
\toprule
\tiny{\texttt{\textit{\underline{Dataset}}:\textbf{iNaturalist-2021}}} \\
\tiny{\texttt{\textit{\underline{Superordinate-level}} \newline
What is the name of the organism that appears in this image? Provide your answer after "Answer:" from one of the following categories: ['Arachnids', 'Mammals', 'Reptiles', 'Animalia', 'Mollusks', 'Plants', 'Amphibians', 'Ray-finned Fishes', 'Birds', 'Insects', 'Fungi']. }} \\
\tiny{\texttt{\textit{\underline{Coarse-level}} \newline
What is the name of the \textcolor{lightblue}{\{concept\_placeholder\}} that appears in this image? For example, if it's a picture of a bengal tiger, give a coarse-grained label for the image 'Tiger'. Provide your answer after "Answer:".}} \\
\tiny{\texttt{\textit{\underline{Fine-level}} \newline
What is the name of the \textcolor{lightblue}{\{concept\_placeholder\}} that appears in this image? For example, if it's a picture of a bengal tiger, give a fine-grained label for the image 'Bengal Tiger' or use its binomial nomenclature 'Panthera tigris tigris'. Provide your answer after "Answer:".}} \\

\cmidrule(lr){1-1}

\tiny{\texttt{\textit{\underline{Dataset}}:\textbf{FGVC-Aircraft}}} \\
\tiny{\texttt{\textit{\underline{Superordinate-level}} \newline
What is the name of the object that appears in this image? Provide your answer after "Answer:" from one of the following categories: ['Airplane', 'Car', 'Train', 'Bicycle', 'Cell Phone', 'Plants', 'Dogs', 'Birds', 'Trucks'].}} \\
\tiny{\texttt{\textit{\underline{Coarse-level}} \newline
What is the manufacturer of the \textcolor{lightblue}{\{concept\_placeholder\}} that appears in this image? Provide your answer after "Answer:" from one of the following categories: ['Embraer', 'Lockheed Corporation', 'Douglas Aircraft Company', 'Cirrus Aircraft', 'Airbus', 'Antonov', 'de Havilland', 'Eurofighter', 'Cessna', 'Tupolev', 'Dornier', 'Yakovlev', 'Panavia', 'Robin', 'ATR', 'Beechcraft', 'Dassault Aviation', 'Fairchild', 'McDonnell Douglas', 'Fokker', 'Gulfstream Aerospace', 'Boeing', 'Saab', 'Canadair', 'Lockheed Martin', 'Supermarine', 'Ilyushin', 'British Aerospace', 'Piper', 'Bombardier Aerospace'].}} \\
\tiny{\texttt{\textit{\underline{Fine-level}} \newline
What is the name of the airplane model made by \textcolor{lightblue}{\{concept\_placeholder\}} that appears in this image? For example, if it's a picture of a Boeing 787 Dreamliner, give a fine-grained label for the image 'Boeing 787 Dreamliner'. Provide your answer after "Answer:".}} \\

\cmidrule(lr){1-1}

\tiny{\texttt{\textit{\underline{Dataset}}:\textbf{Stanford Dogs}}} \\
\tiny{\texttt{\textit{\underline{Superordinate-level}} \newline
What is the name of the organism that appears in this image? Provide your answer after "Answer:" from one of the following categories: ['Arachnids', 'Dogs', 'Reptiles', 'Mollusks', 'Plants', 'Amphibians', 'Ray-finned Fishes', 'Birds', 'Insects', 'Fungi'].}} \\
\tiny{\texttt{\textit{\underline{Coarse-level}} \newline
What is the name of the \textcolor{lightblue}{\{concept\_placeholder\}} that appears in this image? For example, if it's a picture of a Golden Retriever, give a coarse-grained label for the image 'Retriever'.
Provide your answer after "Answer:".}} \\
\tiny{\texttt{\textit{\underline{Fine-level}} \newline
What is the name of the \textcolor{lightblue}{\{concept\_placeholder\}} that appears in this image? For example, if it's a picture of a Golden Retriever, give a coarse-grained label for the image 'Golden Retriever'. Provide your answer after "Answer:".}} \\

\bottomrule
\end{tabular}
\end{adjustbox}
\caption{Prompts for fine-grained image classification. The \textcolor{lightblue}{\{concept\_placeholder\}} is replaced with upper-level concept labels as illustrated in Figure \ref{fig:fgvc_pipeline}.}
\label{table:appendix_fgvc_prompt}
\end{center}
\end{table*}

\begin{table*}[h!]
\addtocounter{table}{-1}
\begin{center}
\begin{adjustbox}{width=1.0\textwidth}
\begin{tabular}{m{12cm}}
\toprule

\tiny{\texttt{\textit{\underline{Dataset}}:\textbf{NABirds}}} \\
\tiny{\texttt{\textit{\underline{Superordinate-level}} \newline
What is the name of the organism that appears in this image? Provide your answer after "Answer:" from one of the following categories: ['Arachnids', 'Mammals', 'Reptiles', 'Animalia', 'Mollusks', 'Plants', 'Amphibians', 'Ray-finned Fishes', 'Birds', 'Insects', 'Fungi'].}} \\
\tiny{\texttt{\textit{\underline{Coarse-level}} \newline
What is the name of the \textcolor{lightblue}{\{concept\_placeholder\}} that appears in this image? For example, if it's a picture of a Owl Parrot, give a coarse-grained label for the image 'Parrot'. Provide your answer after "Answer:".}} \\
\tiny{\texttt{\textit{\underline{Fine-level}} \newline
What is the name of the \textcolor{lightblue}{\{concept\_placeholder\}} that appears in this image? For example, if it's a picture of a Owl Parrot, give a coarse-grained label for the image 'Owl Parrot'. Provide your answer after "Answer:".}} \\
\cmidrule(lr){1-1}

\tiny{\texttt{\textit{\underline{Dataset}}:\textbf{CUB-200-2011}}} \\
\tiny{\texttt{\textit{\underline{Superordinate-level}} \newline
What is the name of the organism that appears in this image? Provide your answer after "Answer:" from one of the following categories: ['Arachnids', 'Mammals', 'Reptiles', 'Animalia', 'Mollusks', 'Plants', 'Amphibians', 'Ray-finned Fishes', 'Birds', 'Insects', 'Fungi'].}} \\
\tiny{\texttt{\textit{\underline{Coarse-level}} \newline
What is the name of the \textcolor{lightblue}{\{concept\_placeholder\}} that appears in this image? For example, if it's a picture of a Owl Parrot, give a coarse-grained label for the image 'Parrot'. Provide your answer after "Answer:".}} \\
\tiny{\texttt{\textit{\underline{Fine-level}} \newline
What is the name of the \textcolor{lightblue}{\{concept\_placeholder\}} that appears in this image? For example, if it's a picture of a Owl Parrot, give a coarse-grained label for the image 'Owl Parrot'. Provide your answer after "Answer:".}} \\

\cmidrule(lr){1-1}

\tiny{\texttt{\textit{\underline{Dataset}}:\textbf{Stanford Cars}}} \\
\tiny{\texttt{\textit{\underline{Superordinate-level}} \newline
What is the name of the object that appears in this image? \
Provide your answer after "Answer:" from one of the following categories: ['Airplane', 'Car', 'Train', 'Bicycle', 'Cell Phone', 'Plants', 'Dogs', 'Birds', 'Trucks'].}} \\
\tiny{\texttt{\textit{\underline{Coarse-level}} \newline
What is the name of the \textcolor{lightblue}{\{concept\_placeholder\}} that appears in this image? Provide your answer after "Answer:" from one of the following categories: ['Sedan', 'SUV', 'Coupe', 'Convertible', 'Pickup', 'Hatchback', 'Van']}} \\
\tiny{\texttt{\textit{\underline{Fine-level}} \newline
What is the name of the \textcolor{lightblue}{\{concept\_placeholder\}} that appears in this image? For example, if it's a picture of a 2006 Honda Civic LX Coupe, give a fine-grained label for the image '2006 Honda Civic LX Coupe'.
Provide your answer after "Answer:".}} \\

\bottomrule
\end{tabular}
\end{adjustbox}
\caption{Prompts for fine-grained image classification. The \textcolor{lightblue}{\{concept\_placeholder\}} is replaced with upper-level concept labels as illustrated in Figure \ref{fig:fgvc_pipeline}.}
\end{center}
\end{table*}

\subsection{Knowledge Probing Prompts}
\label{sec:appendix_knowledge_probe_prompts}
The knowledge probing prompts are shown in Table \ref{table:appendix_knowledge_probe_prompts}. We structure the prompt as explained in Section \ref{sec:probing_concept_knowledge}, where we input the Web-extracted textual attributes along with the coarse-level label for the text-only setting. Since LVLMs are good at identifying the superordinate and coarse-level concepts, we also provide the coarse-level labels as a prior for the text-only setting for a fair comparison in the analysis for fine-grained concept knowledge in the parametric knowledge space.s

\begin{table*}[t]
{\renewcommand{\arraystretch}{1.2}% for the vertical padding
\small
\tabcolsep=0.1cm\centering
\begin{adjustbox}{width=\textwidth}
\begin{tabular}[t]{ lccccccc } 
\toprule
     \shortstack[c]{\textbf{Dataset} \\ \textbf{Name}} & \shortstack[c]{\textbf{Total \# of Instances}\\ \textbf{(Test / Training)}} & \shortstack[c]{\textbf{\# of Superordinate}\\ \textbf{Categories}} & \shortstack[c]{\textbf{\# of Coarse}\\ \textbf{Categories}} & \shortstack[c]{\textbf{\# of Fine}\\ \textbf{Categories}} & \shortstack[c]{\textbf{\# of Images}\\ \textbf{Annotation}} & \shortstack[c]{\textbf{Granularity}\\ \textbf{Hierarchy}} & \shortstack[c]{\textbf{Partial Attribute}\\ \textbf{Annotation}} \\
\midrule
iNaturalist-2021 & 100K / 2.6M & 11 & 1,103 & 10K & 3,286,843 & \checkmark & \texttimes \\
CUB-200-2011 & 5,794 / 5,994 & 1 & 59 & 200 & 11,788 & \texttimes & \checkmark \\
FGVC-Aircraft & 3,333 / 6,667 & 1 & 30 & 100 & 10,000 & \checkmark & \texttimes \\
Stanford Dogs & 8,580 / 12,000 & 1 & - & 120 & 20,580 & \texttimes & \texttimes \\
NABirds & 24,633 / 2,929 & 1 & 146 & 404 & 48,562 & \checkmark & \texttimes \\
Stanford Cars & 8,144 / 8,041 & 1 & 7 & 196 & 16,185 & \texttimes & \texttimes \\
\midrule
    \textbf{\textsc{FINER}} & 372K/2.63M & 16 & 1,416 & 11,171 & 3,393,958 & \checkmark & \checkmark \\
\bottomrule
\end{tabular}
\end{adjustbox}
\caption{\textbf{Overview of the FGVC benchmarks.} Our \textbf{\textsc{Finer}} dataset demonstrates richer set of attributes per concept that enables the evaluation of fine-grained image comprehension. We also augment the benchmarks without \textbf{Granularity Hierarchy} with \textbf{Superordinate Categories} and \textbf{Coarse Categories}.}
\vspace{-0.4cm}
\label{table:data_analysis}
}
\end{table*}

\subsection{Attribute Generation Prompts}
\label{sec:appendix_attr_gen_prompts}
The attribute generation prompts are shown in Table \ref{table:appendix_attribute_generation_prompts}. We divide the attribute generation prompts to three different types: (i) Prompt that generates the Web-extracted attributes given the Wikipedia API retrieved concept documents, (ii) prompt that generates the attributes straight from the model given a text-only input, (iii) prompt that generates the attributes from the model given an image-only input. Note that the prompt variance between the text-only input and image-only input is intentionally minimized, i.e., minimal change in the input prompts, to more accurately isolate the effect of change in the input modalities. The prompts shown in Table \ref{table:appendix_attribute_generation_prompts} are used for the measuring of the modality gap experiments in Section \ref{sec:attribute_generation_modality_gap}.

\subsection{Coarse-Grained Label Generation Prompts}
\label{sec:appendix_coarse_lbl_gen_prompts}
The coarse-grained label generation prompts are shown in Table \ref{table:appendix_coarse_lbl_gen_prompts}. We only generate the coarse-level labels for the following three datasets: (i) Stanford Dogs, (ii) Stanford Cars, (iii) CUB-200-2011, (iv) iNaturalist-2021 because they do not provide concept hierarchy like the rest of the other three datasets. For iNaturalist-2021, although the benchmark does provide the granularity hierarchy, it does so in a taxonomic manner, e.g., order, family, genus, species, which makes it challenging for the model to classify the coarse-grained categories; therefore, we generate coarse-grained labels for the dataset as well. 
By randomly selecting few-shot examples to guide the coarse-grained label generation, we ensure that the generative model, in this case GPT-4V, sticks to the generation of a coarse-grained label. For the faithfulness of the generated coarse-grained labels, we manually evaluate them for datasets other than iNaturalist-2021. For iNaturalist-2021, there are 1,103 coarse-grained categories, which makes it challenging to evaluate each generated coarse-grained label. We therefore group them together with their corresponding \texttt{family}-level category, which serves as a grouping category for the coarse-grained labels. For instance, for \textit{Euphaea fraseri}, we place its coarse-grained labels, \textit{Damselfly} and \textit{Dragonfly} under \textit{Euphaeidae}. By doing so, we not only provide room for more encompassing coarse-grained prediction, e.g., Damselflies are also classified as Dragonflies, but also distinguish the granularity setting from the fine-grained level, which requires a more specific categorization of a given species.

\subsection{Linear Probing Experiment}
\label{sec:appendix_linear_probing}
To evaluate the quality of output representations before and after the multimodal projection in LLaVA-1.5 (7B), we perform linear probing over the output representations of the vision encoder (CLIP-ViT-L/14 \cite{radford2021learning}) and the projected representations in the textual space. The experimental settings are 10 epochs for finetuning and we use accuracy as the evaluation metric. We train on 4x V100 16GB for 57 GPU hours.

\begin{table*}[h!]
\begin{center}
\begin{adjustbox}{width=1.0\textwidth}
\begin{tabular}{m{12cm}}
\toprule

\tiny{\texttt{\textit{\underline{Dataset}}:\textbf{iNaturalist-2021}}} \\
\tiny{\texttt{\textit{\underline{Knowledge Probe Prompt} \newline
Can you guess the specific name (specific epithet) of an organism in the following taxonomic category given its physical attributes?
Provide your answer after "Specific Epithet:".
\newline
Physical Attributes: \textcolor{lightblue}{\{attribute\_placeholder\}} \newline
\newline
Supercategory: \textcolor{lightblue}{\{supercategory\_placeholder\}} \newline
Kingdom: \textcolor{lightblue}{\{kingdom\_placeholder\}} \newline
Phylum: \textcolor{lightblue}{\{phylum\_placeholder\}} \newline
Class: \textcolor{lightblue}{\{class\_placeholder\}} \newline
Order: \textcolor{lightblue}{\{order\_placeholder\}} \newline
Family: \textcolor{lightblue}{\{family\_placeholder\}} \newline
Genus: \textcolor{lightblue}{\{genus\_placeholder\}} \newline
Specific Epithet:
}}} \\

\cmidrule(lr){1-1}

\tiny{\texttt{\textit{\underline{Dataset}}:\textbf{FGVC-Aircraft}}} \\
\tiny{\texttt{\textit{\underline{Knowledge Probe Prompt} \newline
Can you guess the specific name (specific type) of an Airplane in the following taxonomic category given its physical attributes?
Provide your answer after "Specific Airplane:".
\newline
Physical Attributes: \textcolor{lightblue}{\{attribute\_placeholder\}} \newline
\newline
Supercategory: \textcolor{lightblue}{\{supercategory\_placeholder\}} \newline
Coarse-grained Category: \textcolor{lightblue}{\{coarse\_placeholder\}} \newline
Specific Airplane:
}}} 
\\
\cmidrule(lr){1-1}

\tiny{\texttt{\textit{\underline{Dataset}}:\textbf{Stanford Dogs}}} \\
\tiny{\texttt{\textit{\underline{Knowledge Probe Prompt}} \newline
Can you guess the specific name (specific type) of a Dog in the following taxonomic category given its physical attributes?
Provide your answer after "Specific Dog:".
\newline
Physical Attributes: \textcolor{lightblue}{\{attribute\_placeholder\}} \newline
\newline
Supercategory: \textcolor{lightblue}{\{supercategory\_placeholder\}} \newline
Coarse-grained Category: \textcolor{lightblue}{\{coarse\_placeholder\}} \newline
Specific Dog:}} 
\\
\bottomrule
\end{tabular}
\end{adjustbox}
\caption{Prompts for concept-related knowledge probing. The \textcolor{lightblue}{\{concept\_placeholder\}} is replaced with upper-level concept labels as illustrated in Figure \ref{fig:fgvc_pipeline}.}
\label{table:appendix_knowledge_probe_prompts}
\end{center}
\end{table*}

\begin{table*}[h!]
\begin{center}
\begin{adjustbox}{width=1.0\textwidth}
\begin{tabular}{m{12cm}}
\toprule

\tiny{\texttt{\textit{\underline{Attribute Gen. (Text-Only)}} \newline
What are useful visual features for distinguishing \textcolor{lightblue}{\{concept\_placeholder\}} in a photo? Provide the answer as lists of required and likely attributes. For example, for a bengal tiger (Felis Tigris) you might say: \newline
\newline
Required: \newline
- yellow to light orange coat \newline
- dark brown to black stripes \newline
- black rings on the tail \newline
- inner legs and belly are white \newline
- 21 to 29 stripes \newline
Likely: \newline
- lives in mangrove, wooded habitat \newline
- amber, yellow eyes \newline
- large, padded paws \newline
- long tail \newline
- stout teeth \newline
 \newline
 'Required' attributes are a set of external, physical attributes that allows a human to distinguish it from other similar looking concepts. \newline
 'Likely' attributes are a set of attributes that may or may not be visible or are not one of the most discriminative features of the concept. \newline
 In the required (Required:) set, do not include relative, non-visual attributes like size or weight, only the external, visually distinguishable attributes. \newline
 Provide your response in the above format, saying nothing else. If there are no useful visual features, simply write "none". 
}} \\
\tiny{\texttt{\textit{\underline{Attribute Gen. (Wikipedia Doc; Web-Extracted)}} \newline
What are useful visual, external features for distinguishing \textcolor{lightblue}{\{concept\_placeholder\}} in a photo? 
Given an input document (Document:) that may talk about \textcolor{lightblue}{\{concept\_placeholder\}}, provide the answer as lists of required and likely attributes. For example, for a bengal tiger (Felis Tigris) you might say: \newline
\newline
Required: \newline
- yellow to light orange coat \newline
- dark brown to black stripes \newline
- black rings on the tail \newline
- inner legs and belly are white \newline
- 21 to 29 stripes \newline
 \newline
Likely: \newline
- lives in mangrove, wooded habitat \newline
- amber, yellow eyes \newline
- large, padded paws \newline
- long tail \newline
- stout teeth \newline
 \newline
'Required' attributes are a set of external, physical attributes that allows a human to distinguish it from other similar looking concepts. \newline
'Likely' attributes are a set of attributes that may or may not be visible or are not one of the most discriminative features of the concept. \newline
In the required (Required:) set, do not include relative, non-visual attributes like size or weight, only the external, visually distinguishable attributes. \newline
If no document is given, generate from what you already know about \textcolor{lightblue}{\{concept\_placeholder\}}. \newline
Provide your response in the above format, saying nothing else. If there are no useful visual features, simply write "none".
}}
\\
\bottomrule
\end{tabular}
\end{adjustbox}
\caption{Prompts for Attribute Generation. The \textcolor{lightblue}{\{concept\_placeholder\}} is replaced with coarse-grained concept labels for Image-only case and Web-Extracted cases; for Text-only case, use the fine-grained concept label since we want to extract the attributes stored in the parametric knowledge by using the fine-grained concept label as a query.}
\label{table:appendix_attribute_generation_prompts}
\end{center}
\end{table*}

\begin{table*}[ht]
\addtocounter{table}{-1}
\begin{center}
\begin{adjustbox}{width=1.0\textwidth}
\begin{tabular}{m{12cm}}
\toprule

\tiny{\texttt{\textit{\underline{Attribute Gen. (Image-Only)}} \newline
What are useful visual features for distinguishing the \textcolor{lightblue}{\{concept\_placeholder\}} in the photo? 
Provide the answer as lists of required and likely attributes. For example, for a bengal tiger (Felis Tigris) you might say: \newline
\newline
Required: \newline
- yellow to light orange coat \newline
- dark brown to black stripes \newline
- black rings on the tail \newline
- inner legs and belly are white \newline
- 21 to 29 stripes \newline
 \newline
Likely: \newline
- lives in mangrove, wooded habitat \newline
- amber, yellow eyes \newline
- large, padded paws \newline
- long tail \newline
- stout teeth \newline
 \newline
'Required' attributes are a set of external, physical attributes that allows a human to distinguish it from other similar looking concepts. \newline
'Likely' attributes are a set of attributes that may or may not be visible or are not one of the most discriminative features of the concept. \newline
In the required (Required:) set, do not include relative, non-visual attributes like size or weight, only the external, visually distinguishable attributes. \newline
Provide your response in the above format, saying nothing else. If there are no useful visual features, simply write "none".
}} 
\\
\bottomrule
\end{tabular}
\end{adjustbox}
\caption{Prompts for Attribute Generation. The \textcolor{lightblue}{\{concept\_placeholder\}} is replaced with coarse-grained concept labels for Image-only case and Web-Extracted cases; for Text-only case, use the fine-grained concept label since we want to extract the attributes stored in the parametric knowledge by using the fine-grained concept label as a query.}
\end{center}
\end{table*}

\begin{table*}[h!]
\begin{center}
\begin{adjustbox}{width=1.0\textwidth}
\begin{tabular}{m{12cm}}
\toprule

\tiny{\texttt{\textit{\underline{Dataset}}:\textbf{Stanford Cars}  \newline
Generate a coarse-grained label for the following fine-grained car types. \newline
The coarse-grained car types are as follows: ["sedan", "SUV", "coupe", "convertible", "pickup", "hatchback", "van"]. \newline \newline
For example, if the car is a "Ford F-150 Regular Cab 2012" generate "pickup", and if the car is "Chrysler 300 SRT-8 2010", generate "sedan". \newline
Output format is as follows: \newline
\newline
Fine-grained Car Name: Ford F-150 Regular Cab 2012 \newline
Car Name: pickup \newline
\newline
Fine-grained Car Name: Chrysler 300 SRT-8 2010 \newline
Car Name: sedan \newline
\newline
Fine-grained Car Name: Hyundai Santa Fe 2008 \newline
Car Name: SUV
}} \\

\cmidrule(lr){1-1}

\tiny{\texttt{\textit{\underline{Dataset}}:\textbf{CUB-200-2011}  \newline
Generate a coarse-grained label for the following fine-grained bird types.  \newline
For example, if the bird is a "bald eagle (Haliaeetus leucocephalus)" generate "Eagle", and if the bird is "Pine grosbeak", generate "Finch". \newline
Output format is as follows: \newline
\newline
Fine-grained Bird Name: Bald eagle \newline
Bird Name: Eagle \newline
\newline
Fine-grained Bird Name: Pine grosbeak \newline
Bird Name: Finch \newline
\newline
Fine-grained Bird Name: The black backed woodpecker \newline
Bird Name: Woodpecker \newline
}} \\

\cmidrule(lr){1-1}

\tiny{\texttt{\textit{\underline{Dataset}}:\textbf{Stanford Dogs}  \newline
Generate a coarse-grained label for the following fine-grained dog types. \newline
For example, if the dog is a "Cavalier King Charles Spaniel" generate "Spaniel", and if the dog is "The Dear-Headed Chihuahua", generate "Chihuahua". \newline
Output format is as follows: \newline
\newline
Fine-grained Dog Name: Cavalier King Charles Spaniel \newline
Dog Name: Spaniel \newline
\newline
Fine-grained Dog Name: Curly-coated retriever \newline
Dog Name: Retriever \newline
\newline
Fine-grained Dog Name: Newfoundland \newline
Dog Name: Newfoundland \newline
}}
\\
\bottomrule
\end{tabular}
\end{adjustbox}
\caption{Prompts for Coarse-grained Label Generation. We generate the coarse-grained labels for each of the dataset, in case the dataset does not provide the concept hierarchy. The few-shot sampled are randomly sampled from the training instances.}
\label{table:appendix_coarse_lbl_gen_prompts}
\end{center}
\end{table*}

\end{document}